\pdfoutput=1

\documentclass[journal]{IEEEtran}
\ifCLASSINFOpdf
  \usepackage[pdftex]{graphicx}
\else
\fi
\usepackage{amsmath,amssymb}

\usepackage{booktabs}
\usepackage{hyperref}

\usepackage{pifont}
\newcommand{\cmark}{\ding{51}}%
\newcommand{\xmark}{\ding{55}}%

\makeatletter
\newcommand{\figcaption}[1]{\def\@captype{figure}\caption{#1}}
\newcommand{\tblcaption}[1]{\def\@captype{table}\caption{#1}}
\makeatother

\newcommand{\bbR}{\mathbb{R}}

\usepackage[table]{xcolor}

\usepackage{color,soul}

\newtheorem{theorem}{Theorem}
\newtheorem{lemma}{Lemma}

\graphicspath{{figures/}}

\begin{document}
\title{On Data Augmentation for GAN Training}

\author{Ngoc-Trung Tran, 
        Viet-Hung Tran,
        Ngoc-Bao Nguyen,
        Trung-Kien Nguyen,
        Ngai-Man Cheung% <-this % stops a space
%\thanks{Authors were with the Department
%of Electrical and Computer Engineering, Georgia Institute of Technology, Atlanta,
}

\maketitle

\begin{abstract}

Recent successes in Generative Adversarial Networks (GAN) have affirmed the importance of using more data in GAN training. Yet it is expensive to collect data in many domains such as medical applications. Data Augmentation (DA) has been applied in these applications. In this work, we first argue that the classical DA approach could mislead the generator to learn the distribution of the augmented data, which could be different from that of the original data. We then propose a principled framework, termed Data Augmentation Optimized for GAN (DAG), to enable the use of augmented data in GAN training to improve the learning of the {\em original} distribution. We provide theoretical analysis to show that using our proposed DAG aligns with the original GAN in minimizing the Jensen–Shannon (JS) divergence between the {\em  original} distribution and model distribution.  Importantly, the proposed DAG effectively  leverages the augmented data to improve the learning of discriminator and generator. We conduct experiments to apply DAG to different GAN models: unconditional GAN, conditional GAN, self-supervised GAN and CycleGAN using datasets of natural images and medical images. The results show that DAG achieves consistent and considerable improvements across these models. Furthermore, when DAG is used in some GAN models, the system establishes state-of-the-art Fr\'echet Inception Distance (FID) scores. Our code is available\footnote{https://github.com/tntrung/dag-gans}. 

\end{abstract}


%%%%%%%%% BODY TEXT
\section{Introduction}

\IEEEPARstart{G}{enerative} Adversarial Networks  (GANs)~\cite{goodfellow-nisp-2014} is an active research area of generative model learning. GAN has achieved remarkable results in various tasks, for example: image synthesis \cite{karras-iclr-2018,brock-iclr-2018,karras-cvpr-2019,yu-tip-2019ea}, image transformation \cite{isola-cvpr-2017,zhu-cvpr-2017,wang-tip-2018perceptual}, super-resolution \cite{ledig-cvpr-2017,lucas-tip-2019generative}, text to image \cite{reed-arxiv-2016,zhang2-cvpr-2017}, video captioning \cite{yang-tip-2018video}, image dehazing \cite{zhu-ijcai-2018dehazegan}, domain adaptation \cite{zhang-cvpr-2018collaborative}, anomaly detection \cite{schlegl-ipmi-2017,lim-icdm-2018}.
GAN aims to learn the underlying data distribution from a finite number of (high-dimensional) training samples.
The learning is achieved by an adversarial minimax game between a generator $G$ and a discriminator $D$ \cite{goodfellow-nisp-2014}. The minimax game is: 
$\min_G \max_D \mathcal{V}(D,G)$, 
\begin{equation}
\begin{split}
\mathcal{V}(D,G) &=  \mathbb{E}_{\mathbf{x} \sim {P_d}}\log \Big(D(\mathbf{x})\Big)   + \mathbb{E}_{\mathbf{x} \sim {P_g}}\log \Big(1-D(\mathbf{x})\Big)
\end{split}
\label{gan_goodfellow}
\end{equation}
Here, $\mathcal{V}(.)$ is the value function, $P_d$ is the real data distribution of the training samples, $P_g$ is the distribution captured by the generator (G) that maps from the prior noise $\mathbf{z} \sim P_\mathbf{z}$ to the data sample $G(\mathbf{z}) \sim P_g$. $P_\mathbf{z}$ is often Uniform or Gaussian distribution.
It is shown in \cite{goodfellow-nisp-2014}
that given the optimal discriminator $D^*$, $\min_G \mathcal{V}(D^*,G)$ is equivalent to minimizing the Jensen-Shannon (JS) divergence $\mathrm{JS}(P_d||P_g)$.
Therefore, with more samples from $P_d$ (e.g., with a larger training dataset), the empirical estimation of 
$\mathrm{JS}(P_d||P_g)$ can be improved while training a GAN.
This has been demonstrated in recent works \cite{wang-eccv-2018transferring,brock-iclr-2018,donahue-arxiv-2019large}, where GAN benefits dramatically from more data. %in various ways, i.e., bigger data size or more labels.
%scaling the training set and batch-size. 

However, it is widely known that data collection is an extremely expensive process in many domains, e.g. medical images. Therefore, {\em data augmentation}, which has been applied successfully to many deep learning-based discriminative tasks \cite{krizhevsky-nips-2012imagenet,wong-dicta-2016,perez-arxiv-2017}, could be considered for GAN training. In fact, some recent works (e.g. \cite{frid-neurocomputing-2018gan}) have applied label-preserving transformations (e.g. rotation, translation, etc.) to enlarge the training dataset to train a GAN.

However, second thoughts about adding transformed data to the training dataset in training GAN reveal some issues. Some transformed data could be infrequent or non-existence w.r.t. the original data distribution ($P_d(T(\mathbf{x})) \approx 0$, where $T(\mathbf{x})$ is some transformed data by a transformation $T$). On the other hand, augmenting the dataset may mislead the generator to learn to generate these transformed data. For example, if rotation is used for data augmentation on a dataset with category ``horses'', the generator may learn to create rotated horses, which could be inappropriate in some applications.
{\em The fundamental issue is that:  
with data augmentation (DA), the training dataset distribution becomes  $P_d^{\mathcal{T}}$ which could be different from the distribution of the original data $P_d$.  
Following \cite{goodfellow-nisp-2014},
it can be shown that generator learning is minimizing $\mathrm{JS}(P_d^{\mathcal{T}}||P_g)$
instead of $\mathrm{JS}(P_d||P_g)$.
}

{\bf In this work,} 
we conduct a comprehensive study to understand the issue of applying DA for GAN training. The main challenge is to utilize the augmented dataset with distribution $P_d^{\mathcal{T}}$ to improve the learning of $P_d$, distribution of the original dataset. We make the following novel contributions: 

\begin{itemize}
    \item We reveal the issue that the classical way of applying DA for GAN could mislead the generator to create infrequent samples w.r.t. $P_d$.
    \item We propose a new Data Augmentation optimized for GAN (DAG) framework, to leverage augmented samples to improve the learning of GAN to capture \textit{the original distribution}. We discuss {\em invertible transformation} and its JS preserving property. We discuss {\em discriminator regularization} via weight-sharing. We use these as principles to build our framework.
    \item Theoretically, we provide  convergence guarantee of our framework under invertible transformations; empirically, we show that {\em both} invertible and   non-invertible transformations can be used in our  framework to achieve improvement.
    \item We show that our proposed DAG overcomes the issue in classical DA. When DAG is applied to some existing GAN model, we could achieve state-of-the-art performance.
\end{itemize}

\section{Related works}

The standard GAN \cite{goodfellow-nisp-2014} connects the learning of the discriminator and the generator via the single feedback (real or fake) to find the Nash equilibrium in high-dimensional parameter space.  With this feedback, the generator or discriminator may fall into ill-pose settings and get stuck at bad local minimums (i.e. mode collapse) though still satisfying the model constraints. To overcome the problems, different approaches of regularizing models have been proposed.

\textbf{Lipschitzness based Approach.} The most well-known approach is to constrain the discriminator to be 1-Lipschitz. Such GAN relies on methods like weight-clipping \cite{arjovsky-arxiv-2017}, gradient penalty constraints \cite{gulrajani-arxiv-2017,roth-nips-2017,kodali-arxiv-2017,petzka-arxiv-2017,liu-arxiv-2018} and spectral norm \cite{miyato-iclr-2018}. This constraint mitigates gradient vanishing \cite{arjovsky-arxiv-2017} and catastrophic forgetting \cite{tung-icmlw-2018}. However, this approach often suffers the divergence issues \cite{zhang-arxiv-2018,brock-iclr-2018}.

\textbf{Inference Models based Approach.} Inference models enable to infer compact representation of samples, i.e., latent space, to regularize the learning of GAN. For example, using auto-encoder to guide the generator towards resembling realistic samples \cite{makhzani-arxiv-2015}; however, computing reconstruction via auto-encoder often leads to blurry artifacts. VAE/GAN \cite{larsen-arxiv-2015} combines VAE \cite{kingma-arxiv-2013} and GAN, which enables the generator to be regularized via VAE to mitigate mode collapse, and blur to be reduced via the feature-wise distance. ALI \cite{dumoulin-arxiv-2016} and BiGAN \cite{donahue-arxiv-2016} take advantage of the encoder to infer the latent dimensions of the data, and jointly train the data/latent samples in the GAN framework. InfoGAN \cite{chen-arxiv-2016} improves the generator learning via maximizing variational lower bound of the mutual information between the latent and its ensuing generated samples. \cite{tran-eccv-2018,tran-aaai-2018} used auto-encoder to regularize both learning of discriminator and generator. Infomax-GAN \cite{lee-arxiv-2020infomax} applied contrastive learning and mutual information for GAN. It is worth-noting auto-encoder based methods \cite{larsen-arxiv-2015,tran-eccv-2018,tran-aaai-2018}, are likely good to mitigate catastrophic forgetting since the generator is regularized to resemble the real ones. The motivation is similar to EWC \cite{kirkpatrick-2017-nas} or IS \cite{zenke-arxiv-2017}, except the regularization is obtained via the output. Although using feature-wise distance in auto-encoder could reconstruct sharper images, it is still challenging to produce realistic detail of textures or shapes.

\textbf{Multiple Feedbacks based Approach.} The learning via multiple feed-backs has been proposed. Instead of using only one discriminator or generator like standard GAN, the mixture models are proposed, such as multiple discriminators \cite{tu-nips-2017,durugkar-arxiv-2016generative,albuquerque-arxiv-2019multi}, the mixture of generators \cite{hoang-arxiv-2018,ghosh-cvpr-2018} or an attacker applied as a new player for GAN training \cite{liu-cvpr-2019}. \cite{chen-arxiv-2018,tran-arxiv-2019improved,tran-nips-2019} train GAN with auxiliary self-supervised tasks via multi pseudo-classes \cite{gidaris-iclr-2018} that enhance stability of the optimization process.

\textbf{Data-Scale based Approach.} Recent work \cite{brock-iclr-2018,donahue-arxiv-2019large} suggests that GAN benefits from large mini-batch sizes and the larger dataset
\cite{wang-eccv-2018transferring,frid-neurocomputing-2018gan} as many other deep learning models. 
%Larger dataset improves the generalization of learning, while intuitively, with the huge batch size, the probability of one sample appears in the batches is higher that enables GAN to mitigate the catastrophic forgetting more effectively. 
Unfortunately, it is costly to obtain a large-scale collection of samples in many domains. This motivates us to study  Data Augmentation as a potential solution. 
Concurrent with our work,  \cite{zhao-neurips-2020diffaugment,karras-neurips-2020ada,zhao-arxiv-2020image} independently propose data augmentation for training GANs very recently. Our work and all these works are based on different approaches and experiments. 
We recommend the readers to check out their works for more details.
Here, we want to highlight that our work is fundamentally different from these concurrent works:
our framework is based on theoretical JS divergence
preserving  of invertible transformation. Furthermore, we apply the ideas of  multiple discriminators and weight sharing to design our framework. Empirically, we show that both invertible and non-invertible transformations can be used in our framework to achieve improvement especially in setups with  limited data.

%the difference: three of these papers proposed the same augmentation mechanism with a single discriminator and their studies are only with empirical experiments. \textit{Our model is designed with multiple discriminators and works empirically well with most augmentation techniques (both invertible and non-invertible). If the augmentation in use is invertible, the convergence of our model is theoretically guaranteed.}

\section{Notations}

We define some notations to be used in our paper:

\begin{itemize}
    \item $\mathcal{X}$ denotes the original training dataset; $\mathbf{x} \in \mathcal{X}$ has the distribution $P_d$.
    \item $\mathcal{X}^{T}$, $\mathcal{X}^{T_k}$ denote the transformed datasets that are transformed by $T$, $T_k$, resp. 
    $T(\mathbf{x}) \in \mathcal{X}^{T}$ has the distribution $P_d^{T}$; 
    $T_k(\mathbf{x}) \in \mathcal{X}^{T_k}$ has the distribution $P_d^{T_k}$.
    We use $T_1$ to denote an identity transform. Therefore, $\mathcal{X}^{T_1}$ is the original data $\mathcal{X}$. 
    \item $\mathcal{X}^\mathcal{T} =  \mathcal{X}^{T_1} \cup \mathcal{X}^{T_2} \dots \cup \mathcal{X}^{T_K}$ denotes the augmented dataset,  where $\mathcal{T} = \{T_1, T_2, \dots, T_K\}$. Sample in  $\mathcal{X}^\mathcal{T}$
    has the mixture distribution $P_d^{\mathcal{T}}$.
\end{itemize}

\section{Issue of Classical Data Augmentation for GAN}
\label{does_da_works_for_gan}

Data Augmentation (DA) increases the size of the dataset to reduce the over-fitting and generalizes the learning of deep neural networks  \cite{krizhevsky-nips-2012imagenet,wong-dicta-2016,perez-arxiv-2017}. The goal is to improve the classification performance of these networks on the \textit{original dataset}. In this work, we study whether applying DA for GAN can improve learning of the generator and modeling of the distribution $P_d$ of the original dataset. The challenge here is to use additional \textit{augmented data} but have to keep the learning of the \textit{original distribution}. To understand this problem, we first investigate how the classical way of using DA (increasing diversity of $\mathcal{X}$ via transformations $\mathcal{T}$ and use augmented dataset $\mathcal{X}^\mathcal{T}$ as training data for GAN) influences the learning of GAN.

\begin{table*}[t!]
  \caption{The list of DA techniques in our experiments. \cmark: Invertible, \xmark: Non-invertible. Invertible: the original image can be exactly reverted by the inverse transformation. Each original image is transformed into $K-1$ new transformed images. The original image is one class as the identity transformation. FlipRot = Flipping + Rotation.}
  \label{list_of_da}
  \centering
  \begin{tabular}{ccl}
    \toprule
    \textbf{Methods} & \textbf{Invertible} & \textbf{Description}\\
    \midrule
    Rotation            & \cmark & Rotating images with $0^\circ$, $90^\circ$, $180^\circ$ and $270^\circ$ degrees.\\       
    Flipping            & \cmark & Flipping the original image with left-right, bottom-up and the combination of left-right and bottom-up.\\
    Translation         & \xmark & Shifting images $N_t$ pixels in directions: up, down, left and right. Zero-pixels are padded for missing parts caused by\\
                        &        & the shifting.\\
    Cropping            & \xmark & Cropping at four corners with scales $N_c$ of original size and resizing them into the same size as the original image.\\
    FlipRot & \cmark & Combining flipping (left-right, bottom-up) + rotation of $90^\circ$.\\
    \bottomrule
  \end{tabular}
  \label{da_techniques}
\end{table*}

\begin{figure}
  \centering
    \includegraphics[width=4.3cm,keepaspectratio]{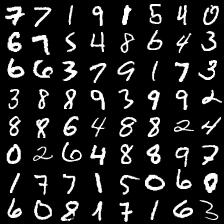}
    \includegraphics[width=4.3cm,keepaspectratio]{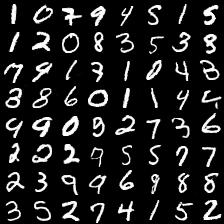}
    
    \vspace{0.1cm}
    
    \includegraphics[width=4.3cm,keepaspectratio]{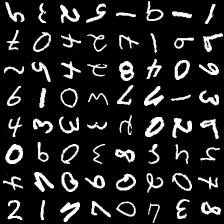}
    \includegraphics[width=4.3cm,keepaspectratio]{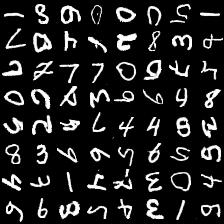}
  \caption{Generated examples of the full MNIST dataset (100\%). Top row: the real samples, the generated samples of Baseline model. Bottom row: the rotated real samples and the generated samples of DA with rotation. %Refer to Fig. \ref{mnist_data_agumentation_others} in Appendix \ref{appendix_b} for other examples of flipping and cropping.
  }
  \label{mnist_data_agumentation}
\end{figure}

\begin{equation}
\mathcal{V}(D,G) =  \mathbb{E}_{\mathbf{x} \sim {P_d^\mathcal{T}}}\log \Big(D(\mathbf{x})\Big)   + \mathbb{E}_{\mathbf{x} \sim {P_g}}\log \Big(1-D(\mathbf{x})\Big)
\label{gan_original}
\end{equation}

\noindent \textbf{Toy example.} We set up the toy example with the MNIST dataset for the illustration. In this experiment, we augment the original MNIST dataset (distribution $P_d$) with some widely-used augmentation techniques $\mathcal{T}$ (rotation, flipping, and cropping) (Refer to Table. \ref{da_techniques} for details) to obtain new dataset (distribution $P_d^\mathcal{T}$). Then, we train the standard GAN \cite{goodfellow-nisp-2014} (objectives is shown in Eq. \ref{gan_original}) on this new dataset. We construct two datasets with two different sizes:  \textbf{100\%} and \textbf{25\%} randomly selected MNIST samples. We denote the GAN model trained on the original dataset as \textbf{Baseline}, and GAN trained on the augmented dataset as \textbf{DA}. We evaluate models by FID scores. We train the model with 200K iterations using small DCGAN architecture similar to \cite{gulrajani-arxiv-2017}. We compute the 10K-10K FID \cite{heusel-arxiv-2017} (using a pre-trained MNIST classifier) to measure the similarity between the generator distributions and the distribution of the original dataset.  For a fair comparison, we use K = 4 for all augmentation methods.

\begin{table}
    \centering
    %\footnotesize
    \caption{Best FID (10K-10K) of GAN baseline with classical DA on MNIST dataset.}
    \begin{tabular}{c c c c c c c c}
    \toprule
    \textbf{Data size} & \textbf{Baseline}  & \textbf{Rotation} & \textbf{Flipping}  & \textbf{Cropping}\\
    \midrule
    \textbf{100\%} & 6.8 & 73.1 & 47.3  & 114.4 \\
    \textbf{25\%}  & 7.5 & 72.5 & 46.2  & 114.2 \\
    %\textbf{10\%}  & 10.2 & - & - & - & - & - & - &  - & -  \\
    \bottomrule
    \end{tabular}
    \label{mnist_fid}
\end{table}

Some generated examples of Baseline and DA methods are visualized in Fig. \ref{mnist_data_agumentation}. Top row: real samples, the generated samples of the Baseline model. Bottom row: the rotated real samples and the generated samples of DA with rotation. See more examples of DA with flipping and cropping in Fig. \ref{mnist_data_agumentation_others} of Appendix \ref{appendix_b} in the supplementary material. We observe that the generators trained with DA methods create samples similar to the augmented distribution $P_d^\mathcal{T}$. Therefore, many generated examples are out of $P_d$. To be precise, we measure the similarity between the generator distribution $P_g$ and $P_d$ with FID scores as in Table. \ref{mnist_fid}. The FIDs of DA methods are much higher as compared to that of Baseline for both cases 100\% and 25\% of the dataset. 
This suggests that 
applying DA in the 
classical way would misguide the generator to learn a rather different distribution compared to that of the original data.  Comparing different augmentation techniques, it makes sense that the distributions of DA with flipping and DA with cropping are most similar and different from the original distribution respectively. Training DA on small/full dataset results in FID difference for Baseline. It means there are some impacts of data size on the learning of GAN (to be discussed further). We further support these observations with the theoretical analysis in Sec. \ref{theory_analysis_da}. This experiment illustrates that applying DA in a classical way for GAN could  encounter an issue:  infrequent  samples may be generated more due to alternation in the data distribution $P_d$. Therefore, the classical way of applying DA may not be suitable  for GAN. To apply data augmentation in GAN,  the methods of applying DA need to ensure the learning of $P_d$. We propose a new DA framework to achieve this.

\subsection{Theoretical Analysis on DA}
\label{theory_analysis_da}

Generally, let $\mathcal{T} = \{T_1, T_2, \dots, T_K\}$ be the set of augmentation techniques to apply on the original dataset. $P_d$ is the distribution of original dataset. $P_d^\mathcal{T}$ is the distribution of the augmented dataset. Training GAN \cite{goodfellow-nisp-2014} on this new dataset, the generator is trained via minimizing the JS divergence between its distribution $P_g$ and $P_d^\mathcal{T}$ as following (The proof is similar in \cite{goodfellow-nisp-2014}).

\begin{equation}
\mathcal{V}(D^*,G) = -\log(4) + 2 \cdot \mathrm{JS}(P_d^\mathcal{T}||P_g)
\end{equation}

where $D^*$ is the optimal discriminator. Assume that the optimal solution can be obtained: $P_g = P_d^\mathcal{T}$.

\section{Proposed method}
\label{proposed_method}

The previous section illustrates the issue of classical \textbf{DA} for GAN training. The challenge here is to use the augmented dataset $\mathcal{X}^\mathcal{T}$ to improve the learning of the distribution of {\em original data}, i.e. $P_d$ instead of $P_d^{\mathcal{T}}$.
To address this,

\begin{enumerate}
    \item We first discuss {\em invertible transformations} and their {\em invariance} for JS divergence.
    \item We then present a simple modification of the 
vanilla GAN that is capable to learn $P_d$ using {\em transformed} samples $\mathcal{X}^{T_k}$, provided that the transformation is invertible as discussed in (1).
    \item Finally, we present our model which is a stack of the modified GAN in (2); we show that this model is capable to use the augmented dataset $\mathcal{X}^\mathcal{T}$,
    where $\mathcal{T} = \{T_1, T_2, \dots, T_K\}$,
    to improve the learning of $P_d$. 
\end{enumerate}

\subsection{Jensen-Shannon (JS) Preserving with Invertible Transformation}
\noindent \textbf{Invertible mapping function \cite{qiao-tip-2010study}.} Considering two distributions $p_\mathbf{x}(\mathbf{x})$ and $q_\mathbf{x}(\mathbf{x})$ in space $\mathbb{X}$. Let $T$: $\mathbb{X} \rightarrow \mathbb{Y}$ denote the differentiable and invertible (bijective) mapping function (linear or non-linear) that converts $\mathbf{x}$ into $\mathbf{y}$, i.e. $\mathbf{y} = T(\mathbf{x})$. Then we have the following theorem:

\begin{theorem}
The Jensen-Shannon (JS) divergence between two distributions is invariant under differentiable and invertible transformation $T$:
\begin{equation}
\mathrm{JS}(p_\mathbf{x}(\mathbf{x})||q_\mathbf{x}(\mathbf{x})) = \mathrm{JS}(p_\mathbf{y}(\mathbf{y})||q_\mathbf{y}(\mathbf{y}))
\end{equation}
\label{js_theorem_1}
\end{theorem}

\noindent \textit{Proof}. Refer to our proof in Appendix \ref{proofs_for_theorems}. 
In our case, we have $p_\mathbf{x}(.), q_\mathbf{x}(.),  p_\mathbf{y}(.), q_\mathbf{y}(.)$ to be $P_d, P_g, P_d^T, P_g^T$ resp.
Thus, if an invertible transformation is used, then 
$\mathrm{JS}(P_d || P_g)=\mathrm{JS}(P_d^T || P_g^T)$.
Note that, 
if $T$ is non-invertible,  
$\mathrm{JS}(P_d^T || P_g^T)$ may approximate $\mathrm{JS}(P_d || P_g)$ to some extent. The detailed investigation of this situation is beyond the scope of our work.
However, the take-away from this theorem is that {\em JS preserving can be guaranteed if invertible transformation is used}.

% \begin{theorem}
% Similar to Theorem \ref{js_theorem_1}. Let $T$ be arbitrary the transformation, we have: $\mathrm{JS}(P_g^{T}||P_d^{T}) \leq \mathrm{JS}(P_g||P_d)$.
% \label{js_theorem_2}
% \end{theorem}

% \noindent \textit{Proof}. Refer to our proof in Appendix \ref{appendix_a}.\\

\subsection{GAN Training with Transformed Samples}
\label{gan_trainining_with_transformed_samples}

\begin{figure*}
    \centering
    \includegraphics[width=8.0cm,keepaspectratio]{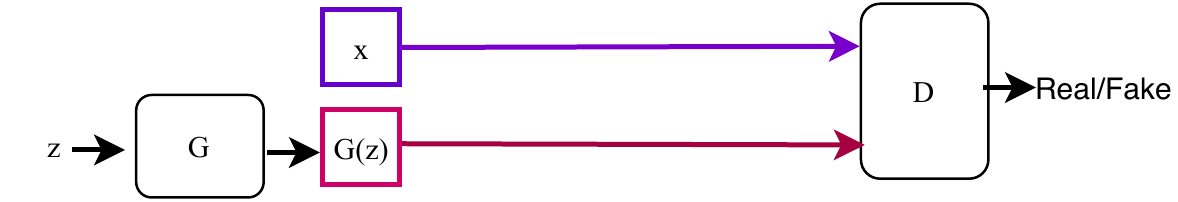} 
    \includegraphics[width=8.0cm,keepaspectratio]{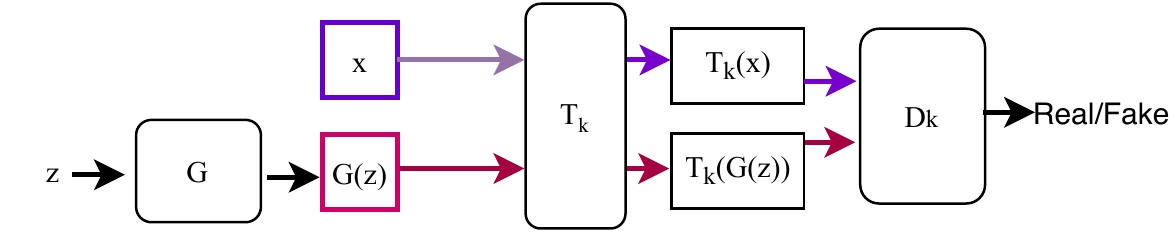}
    \caption{The original (vanilla) GAN model (left) and our design to train GAN with transformed data (right).}
    \label{gan_models}
\end{figure*}

Motivated by this invariant property of JS divergence, we design the GAN training mechanism to utilize the transformed data, but still, preserve the learning of $P_d$ by the generator.
Figure~\ref{gan_models}
illustrates the vanilla GAN (left) and this new design (right).
Compared to the vanilla GAN, the change is simple: the real and fake samples are transformed by $T_k$ before feeding into the discriminator $D_k$. 
Importantly, 
generator's samples are {\em transformed} to imitate the transformed real samples, thus the generator is guided to learn the distribution of the original data samples in $\mathcal{X}$. The mini-max objective of this design is same as that of the vanilla GAN, except that now the discriminator sees the transformed real/fake samples:

\begin{equation}
\mathcal{V}(D_k, G) =\mathbb{E}_{\mathbf{x} \sim {P_d^{T_k}}}\log \Big(D_k(\mathbf{x})\Big)
+ \mathbb{E}_{\mathbf{x} \sim {P_g^{T_k}}}\log \Big(1-D_k(\mathbf{x})\Big)  
\label{eq:gan_transformed_sample}
\end{equation}

\begin{figure*}
    \centering
    \includegraphics[width=8.0cm,keepaspectratio]{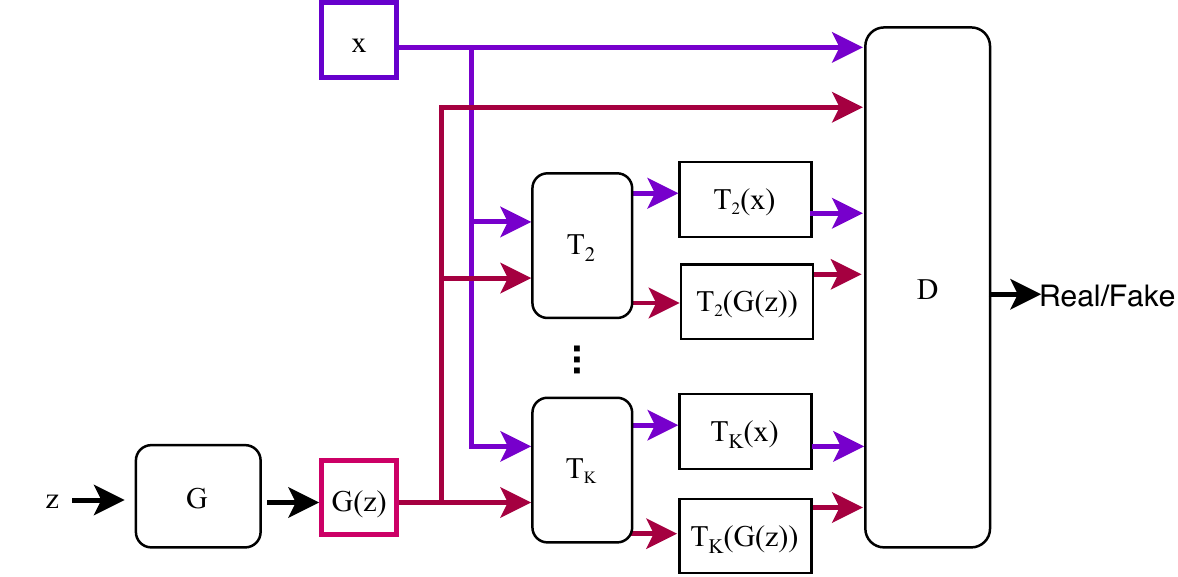} 
    \includegraphics[width=8.0cm,keepaspectratio]{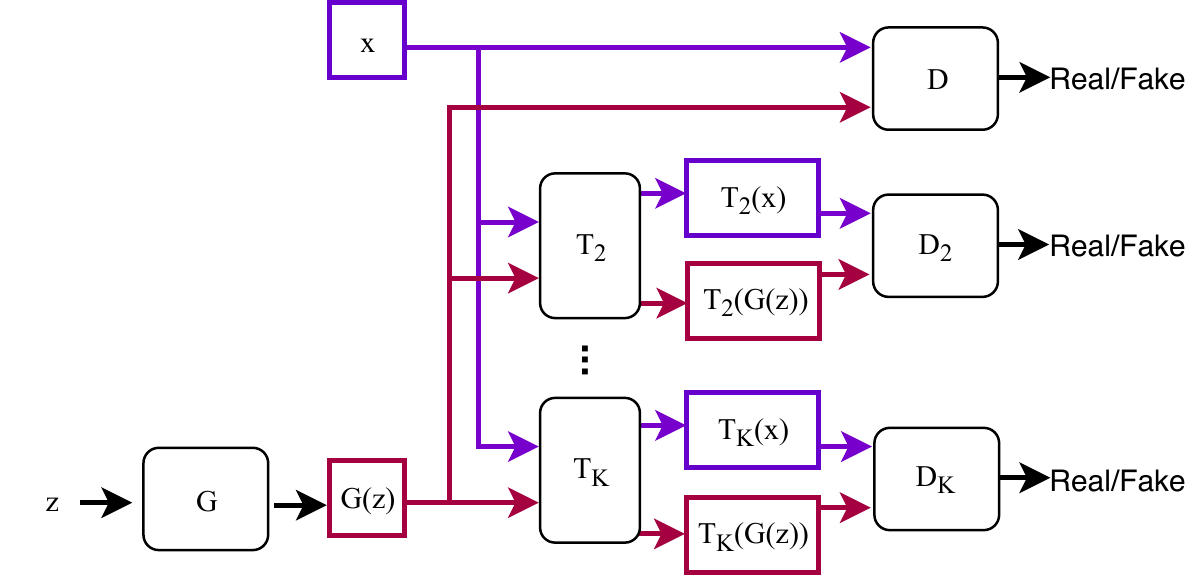}
    \caption{(a) IDA model with single discriminator (b) Our final proposed model (DAG) with multiple discriminators $D_k$. In these models, both real samples (violet paths) and fake samples (red paths) are used to train discriminators. Only fake samples (red paths) are used to train the generator.}
    \label{dag_models}
\end{figure*}

where $P_d^{T_k}, P_g^{T_k}$ be the distributions of {\em transformed} real and fake data samples respectively, $T_k \in \mathcal{T}$. For fixed generator $G$, the optimal discriminator $D^*_k$ of $\mathcal{V}(D_k,G)$ is that in Eq. \ref{optimal_c_2} (the proof follows the arguments as in \cite{goodfellow-nisp-2014}). With the \textit{invertible transformation} $T_k$, $D_k$ is trained to achieve exactly \textit{the same optimal} as $D$:

\begin{equation}
\begin{split}
D_k^*(T_k(\mathbf{x})) 
&= \frac{p_d^{T_k}(T_k(\mathbf{x}))}{p_d^{T_k}(T_k(\mathbf{x})) + p_g^{T_k}(T_k(\mathbf{x}))} \\
& =
\frac{p_d(\mathbf{x})|\mathcal{J}^{T_k}(\mathbf{x})|^{-1}}{p_d(\mathbf{x})|\mathcal{J}^{T_k}(\mathbf{x})|^{-1} + p_g (\mathbf{x}) |\mathcal{J}^{T_k}(\mathbf{x})|^{-1}} \\
&= \frac{p_d(\mathbf{x})}{p_d(\mathbf{x}) + p_g(\mathbf{x})} = D^*(\mathbf{x})
\end{split}
\label{optimal_c_2}
\end{equation}
where $|\mathcal{J}^{T_k}(\mathbf{x})|$ is the determinant of Jacobian matrix of $T_k$. Given optimal $D_k^*$, training generator with these transformed samples is equivalent to minimizing JS divergence between $P_d^{T_k}$ and $P_g^{T_k}$:

\begin{equation}
\begin{split}
\mathcal{V}(D_k^*,G) &= -\log(4) + 2\cdot \mathrm{JS}(P_d^{T_k}||P_g^{T_k})
\end{split}
\end{equation}

Furthermore, if an invertible transformation is chosen for $T_k$, then
$\mathcal{V}(D_k^*,G) = -\log(4) + 2\cdot \mathrm{JS}(P_d||P_g)$ (using Theorem 1).
Therefore, 
this mechanism guarantees the generator to learn to create the original samples, not transformed samples. The convergence of GAN with transformed samples has the same JS divergence as the original GAN if the transformation $T_k$ is invertible. 
Note that, this design  has no advantage over the original GAN: it performs the same as the original GAN. However, we explore a design to stack them together to utilize augmented samples with multiple transformations. This will be discussed next.

\subsection{Data Augmentation Optimized for GAN}
\label{da_minimax_gan_ss}

{\bf Improved DA (IDA).}
Building on the design of the previous section, 
we make the first attempt to 
leverage the augmented samples for GAN training as shown in Fig. \ref{dag_models}a, termed Improved DA (IDA).
Specifically, we transform fake and real samples with $\{T_k\}$ and feed the mixture of those transformed real/fake samples as inputs to train a \textit{single} discriminator D (recall that $T_1$ denotes the identity transform). Training the discriminator (regarded as a binary classifier) using augmented samples tends to improve generalization of discriminator learning: i.e., by increasing feature invariance (regarding real vs. fake) to specific transformations, and penalizing model complexity via a regularization term based on the variance of the augmented forms \cite{dao-pmlr-2019kernel}. Improving feature representation learning is important to improve the performance of GAN \cite{chen-nips-2016,chen-arxiv-2018}. However, although IDA can benefit from
invertible transformation as in Section \ref{gan_trainining_with_transformed_samples}, training all samples with a \textit{single} discriminator {\em does not} preserve JS divergence of original GAN (Refer to Theorem \ref{theorem_2} of Appendix \ref{appendix_a} for proofs). IDA is our first attempt to use augmented data to improve GAN training. Since it is not JS preserving, it does not guarantee the convergence of GAN. We state the issue of IDA  in Theorem \ref{theorem_2}.

\begin{theorem}
Considering two distributions $p, q$: $p = \sum_{m=1}^{K}w_{m}p^m$ and $q = \sum_{m=1}^{K}w_{m}q^m$, where $\sum_{m=1}^{K}w_{m} = 1$. If distributions $p^m$ and $q^m$ are distributions of $p^0$ and $q^0$ transformed by invertible transformations $T_m$ respectively, we have:
\begin{equation}
\mathrm{JS}(p||q) \leq \mathrm{JS}(p^0||q^0)    
\end{equation}
\label{theorem_2}
\end{theorem}

\noindent \textit{Proofs.} From Theorem \ref{js_theorem_1} of invertible transformation $T_m$, we have $\mathrm{JS}(p^0||q^0) = \mathrm{JS}(p^m||q^m)$. Substituting this into the Lemma \ref{lemma_2} (in Appendix): $\mathrm{JS}(p||q)\leq\sum_{m=1}^{K}w_{m}\mathrm{JS}(p^0||q^0) = (\sum_{m=1}^{K}w_{m})\mathrm{JS}(p^0||q^0) = \mathrm{JS}(p^0||q^0)$. It concludes the proof.\\

\noindent In our case, we assume that $p, q$ are mixtures of distributions that are inputs of IDA method (discussed in Section \ref{da_minimax_gan_ss}) and $p^0 = P_d$, $q^0 = P_g$. In fact, the mixture of transformed samples has the form of distributions as discussed in Theorem \ref{theorem_2} (Refer to Lemma \ref{lemma_1} in Appendix). According to Theorem \ref{theorem_2}, IDA method is minimizing the lower-bound of JS divergence instead of the exact divergence of $\mathrm{JS}(P_d||P_g)$. Due to this issue, although using more augmented samples, but IDA (FID = 29.7) does not out-perform the Baseline (FID = 29.6) (Refer to Table \ref{distgan_on_shared_weights} in Section \ref{experiments}).

{\bf Data Augmentation optimized for GAN (DAG).}
In what follows, we discuss another proposed framework to overcome the above problem.
The proposed  Data Augmentation optimized for GAN (DAG) aims to utilize an augmented dataset $\mathcal{X}^\mathcal{T}$ with samples transformed by $\mathcal{T} = \{T_1, T_2, \dots, T_K\}$ to improve learning of the distribution of {\em original data} (Fig. \ref{dag_models}b). DAG takes advantage of the different transformed samples by using different discriminators $D, \{D_k\} = \{D_2, D_3, \dots D_K\}$. The discriminator $D_k$ is trained on samples \textit{transformed} by $T_k$.

\begin{equation}
\begin{split}
\max_{D,\{D_k\}}&\mathcal{V}(D,\{D_k\},G) = \mathcal{V}(D,G) + \frac{\lambda_u}{K-1}\sum_{k=2}^{K} \mathcal{V}(D_k,G)
\end{split}
\label{gan_dis_aux}
\end{equation}

We form our discriminator objective by augmenting the original GAN discriminator objective 
$\mathcal{V}(D,G)$
with $\mathcal{V}(\{D_k\},G) = \sum_{k=2}^{K}\mathcal{V}(D_k,G)$, see Eq. \ref{gan_dis_aux}. Each objective $\mathcal{V}(D_k,G)$ is
given by Eq. 
\ref{eq:gan_transformed_sample}, i.e., 
similar to original GAN objective \cite{goodfellow-nisp-2014} except that the inputs to discriminator are now transformed, as discussed previously. $D_k$ is trained to distinguish transformed real samples vs. transformed fake samples (both transformed by same $T_k$).

\begin{equation}
\begin{split}
\min_G\mathcal{V}(D,\{D_k\},G) &= \mathcal{V}(D,G) + \frac{\lambda_v}{K-1} \sum_{k=2}^{K} \mathcal{V}(D_k,G)
\end{split}
\label{gan_gen_aux}
\end{equation}

Our generator objective is shown in Eq. \ref{gan_gen_aux}. The generator $G$ learns to create samples to fool the discriminators $D$ and $\{D_k\}$ {\em simultaneously}. The generator takes the random noise $\mathbf{z}$ as input and maps into  $G(\mathbf{z})$ to confuse $D$ as in standard GAN. It is important as we want the generator to generate only original images, not transformed images. Then, $G(\mathbf{z})$ is transformed by $T_k$ to confuse $D_k$ in the corresponding task $\mathcal{V}(D_k, G)$. Here, $\mathcal{V}(\{D_k\},G) = \sum_{k=2}^K \mathcal{V}(D_k,G)$. 
When leveraging the transformed samples, the generator receives $K$ feed-back signals to learn and improve itself in the adversarial mini-max game.
If the generator wants its created samples to look realistic, the transformed counterparts need to look realistic also. The feedbacks are computed from not only $\mathrm{JS}(P_d||P_g)$ of the original samples but also $\mathrm{JS}(P_d^{T_k}||P_g^{T_k})$ of the transformed samples as discussed in the next section. In Eq. \ref{gan_dis_aux} and \ref{gan_gen_aux}, $\lambda_u$ and $\lambda_v$ are constants.

\subsubsection{Analysis on JS preserving}
\label{js_analysis_preserving}

The invertible transformations ensure no discrepancy in the optimal convergence of discriminators, i.e., $D_k$ are trained to achieve \textit{the same optimal} as $D$: $D_k^*(T_k(\mathbf{x})) = D^*(\mathbf{x}), \forall k$ (Refer to Eq. \ref{optimal_c_2}). Given these optimal discriminators $\{D_k^*\}$ at equilibrium point. For generator learning, minimizing $\mathcal{V}(\{D_k\},G)$ in Eq. \ref{gan_gen_aux} is equivalent to minimizing Eq. \ref{g_obj_js}:

\begin{equation}
\begin{split}
\mathcal{V}(\{D_k^*\},G) &= \mathrm{const} + 2 \sum_{k=2}^{K}\mathrm{JS}(P_d^{T_k}||P_g^{T_k})
\end{split}
\label{g_obj_js}
\end{equation}

Furthermore, if all $T_k$ are invertible, the r.h.s. of 
Eq. \ref{g_obj_js} becomes:
$ \mathrm{const} + 2 (K-1) \cdot \mathrm{JS}(P_d||P_g)$. In this case, the convergence of GAN is guaranteed. 
In this attempt, $D, \{D_k\}$ 
do not have any shared weights. 
Refer to Table \ref{distgan_on_shared_weights} in Section \ref{experiments}: when we use multiple discriminators $D, \{D_k\}$ to handle transformed samples by $\{T_k\}$ respectively, the performance is slightly improved to FID = 28.6 (``None" DAG) from Baseline (FID = 29.6). This verifies the advantage of JS preserving of our model in generator learning. 
To further improve the design, 
we propose to apply weight sharing for  $D, \{D_k\}$, so that we can take  advantage of data augmentation, i.e. via improving feature representation learning of discriminators.

\subsubsection{Discriminator regularization via weight sharing}

We propose to regularize the learning of discriminators by enforcing weights sharing between them. Like IDA, \textit{discriminator gets benefit from the data augmentation to improve the representation learning of discriminator and furthermore, the model preserves the same JS objective to ensure the convergence of the original GAN.} Note that the number of shared layers between discriminators does not influence the JS preserving property in our DAG (the same proofs about JS as in Section \ref{gan_trainining_with_transformed_samples}). The effect of number of shared layers will be examined via experiments (i.e., in Table \ref{distgan_on_shared_weights} in Section \ref{experiments}). Here, we highlight that with discriminator regularization (on top of JS preserving), the performance is substantially improved. In practical implementation, $D, \{D_k\}$ shared all layers except the last layers to implement different heads for different outputs. See Table \ref{distgan_on_shared_weights} in Section \ref{experiments} for more details. %We also discuss the perspective of DAG in GAN training in Appendix \ref{healthy_interaction}.

%The improvement of GAN training is clear: the original GAN is augmented with multiple branches that work together to minimize the same JS divergence between $P_d$ and $P_g$. In addition, it enables to use of augmented samples to improve the feature representation in discriminator learning via weight-sharing while still preserving the convergence of the original generator.

In this work, we focus on invertible transformation in image domains. In the image domain, the transformation is invertible if its transformed sample can be reverted to the exact original image. For example, some popular affine transformations in image domain are rotation, flipping or fliprot (flipping + rotation), etc.; However, empirically, we find out that our DAG framework works favorably with most of the augmentation techniques (even non-invertible transformation) i.e., cropping and translation. However, if the transformation is invertible, the convergence property of GAN is theoretically guaranteed. Table \ref{da_techniques} represents some examples of invertible and non-invertible transformations that we study in this work.

The usage of DAG outperforms the baseline GAN models (refer to Section \ref{exp_data_augmentation} for details). Our DAG framework can apply to various  GAN models: unconditional GAN, conditional GAN, self-supervised GAN, CycleGAN.
Specifically, the same ideas can be applied: train the discriminator with  transformed real/fake samples as inputs; train the generator with 
transformed fake samples to learn $P_d$; stack such modified models and apply weight sharing to leverage different transformed samples.
We select one state-of-the-art GAN system recently published \cite{tran-nips-2019} and apply DAG. We refer to this as our best GAN system; this system advances state-of-the-art performance on benchmark datasets, as will be discussed next.

%Like invertible transformation, we think there are also theoretical reasons behind for non-invertible transformations like minimizing the upper-bound of JS divergence (i.e., similar to Theorem 4 of \cite{qiao-tip-2010study}) instead of exact JS divergence of the original distribution. We let it as future work. 
%In this work, we focus on invertible transformations and provide the empirical experiments for those non-invertible transformations to prove the effectiveness of our proposed DAG on various augmentation techniques.

\subsubsection{Difference from existing works with multiple discriminators} We highlight the difference between our work and existing works that also uses multiple discriminators \cite{tu-nips-2017,durugkar-arxiv-2016generative,albuquerque-arxiv-2019multi}: i) we use augmented data to train multiple discriminators, ii) we propose the DAG architecture with invertible transformations that preserve the JS divergence as the original GAN. Furthermore, our DAG is simple to implement on top of any GAN models and potentially has no limits of augmentation techniques or number discriminators to some extent. Empirically, the more augmented data DAG uses (adhesive to the higher number of discriminators), the better FID scores it gets.

\section{Experiments}
\label{experiments}

We first conduct the ablation study on DAG, then investigate the influence of DAG across various augmentation techniques on two state-of-the-art baseline models:   
Dist-GAN \cite{tran-eccv-2018} for unconditional GAN and 
SS-GAN \cite{chen-arxiv-2018} for self-supervised GAN.
Then, we introduce our best system by making use of DAG on top of a recent GAN system to compare to the state of the art. %See the details of model training parameters and experimental setup in Appendix \ref{model_training}.

\textbf{Model training.} We use batch size of 64 and the latent dimension of $d_\mathrm{z} = 128$ in most of our experiments (except in Stacked MNIST dataset, we have to follow the latent dimension as in \cite{metz-arxiv-2016}). We train models using Adam optimizer with learning rate $\mathrm{lr} = 2 \times 10^{-4}$, $\beta_1 = 0.5$, $\beta_2 = 0.9$ for DCGAN backbone \cite{radford-arxiv-2015} and $\beta_1 = 0.0$, $\beta_2 = 0.9$ for Residual Network (ResNet) backbone \cite{gulrajani-arxiv-2017}. We use linear decay over 300K iterations for ResNet backbone as in \cite{gulrajani-arxiv-2017}. We use our best parameters: $\lambda_u = 0.2$, $\lambda_v = 0.2$ for SS-GAN and $\lambda_u = 0.2$, $\lambda_v = 0.02$ for Dist-GAN. We follow \cite{chen-arxiv-2018} to train the discriminator with two critics to obtain the best performance for SS-GAN baseline. For fairness, we implement DAG with $K = 4$ branches for all augmentation techniques, and the number of samples in each training batch are equal for DA and DAG. In our implementation, $N_t = 5$ pixels for translation and the cropping scale $N_c = 0.75$ for cropping (Table \ref{da_techniques}).

\textbf{Evaluation.} We perform extensive experiments on datasets: CIFAR-10, STL-10, and Stacked MNIST. We measure the diversity/quality of generated samples via FID \cite{heusel-arxiv-2017} for CIFAR-10 and SLT-10. FID is computed with 10K real samples and 5K generated samples as in \cite{miyato-iclr-2018} if not precisely mentioned. We report the best FID attained in 300K iterations as in \cite{xiang-arxiv-2017,li-nips-2017,tran-eccv-2018,yazici-arxiv-2018}. In FID figures, The horizontal axis is the number of training iterations, and the vertical axis is the FID score. We report the number of modes covered (\#modes) and the KL divergence score on Stacked MNIST similar to \cite{metz-arxiv-2016}.

\subsection{Ablation study}
\label{ablation_study}

We conduct the experiments to verify the importance of discriminator regularization, and JS preserving our proposed DAG. In this study, we mainly use Dist-GAN and SS-GAN as baselines and train on full (100\%) CIFAR-10 dataset. For DAG, we use K = 4 rotations. As the study requires expensive computation, we prefer the small DC-GAN network (Refer to Appendix \ref{appendix_d} for details). The network backbone has four conv-layers and 1 fully-connect (FC) layer.

\subsubsection{The impacts of discriminator regularization}
We validate the importance of discriminator regularization (via shared weights) in DAG. We compare four variants of DAG: i) discriminators share no layers (None), ii) discriminators share a half number of conv-layers (Half), which is two conv-layers in current model, iii) discriminators share all layers (All), iv) discriminators share all layers but FC (All but heads). Note that the number of shared layers counts from the first layer of the discriminator network in this study. As shown in Table \ref{distgan_on_shared_weights}, comparing to Baseline, DA, and IDA, we can see the impacts of shared weights in our DAG. \textit{This verifies the importance of discriminator regularization in DAG.} In this experiment, two settings: ``Half" and ``All but heads", achieve almost similar performance, but the latter is more memory-efficient, cheap and consistent to implement in any network configurations. Therefore, we choose ``All but heads" for our DAG setting for the next experiments. Dist-GAN is the baseline for this experiment. 
%man: comment this since the difference is not significant
%Note that ``All"-DAG and IDA are quite similar and have the same number of parameters, but thanks to JS preserving, ``All"-DAG significantly outperforms IDA.

\begin{table*}
  \caption{The ablation study on discriminator regularization via the number of shared layers (counts from the first layer) in our DAG model. "None": sharing no layers at all. "Half": sharing a half number of layers of the network. "All but heads": sharing all convolutional layers but different FC layers. "All": sharing all layers. Baseline: Dist-GAN.}
  \label{ablation_study_md_distgan}
  \centering
  \begin{tabular}{l|c|c|c|c||c|c|c}
    \toprule
    \textbf{Shared layers} & None & Half & All but heads & All & Baseline & DA & IDA   \\
    \hline
    \textbf{FID} & 28.6 & 23.9 & \textbf{23.7} & 26.0 & 29.6 & 49.0 & 29.7  \\
    \bottomrule
  \end{tabular}
  \label{distgan_on_shared_weights}
\end{table*}

\subsubsection{The importance of JS preserving and the role of transformations in generator learning of DAG}

\begin{figure}
    \centering
    \includegraphics[width=8.0cm,keepaspectratio]{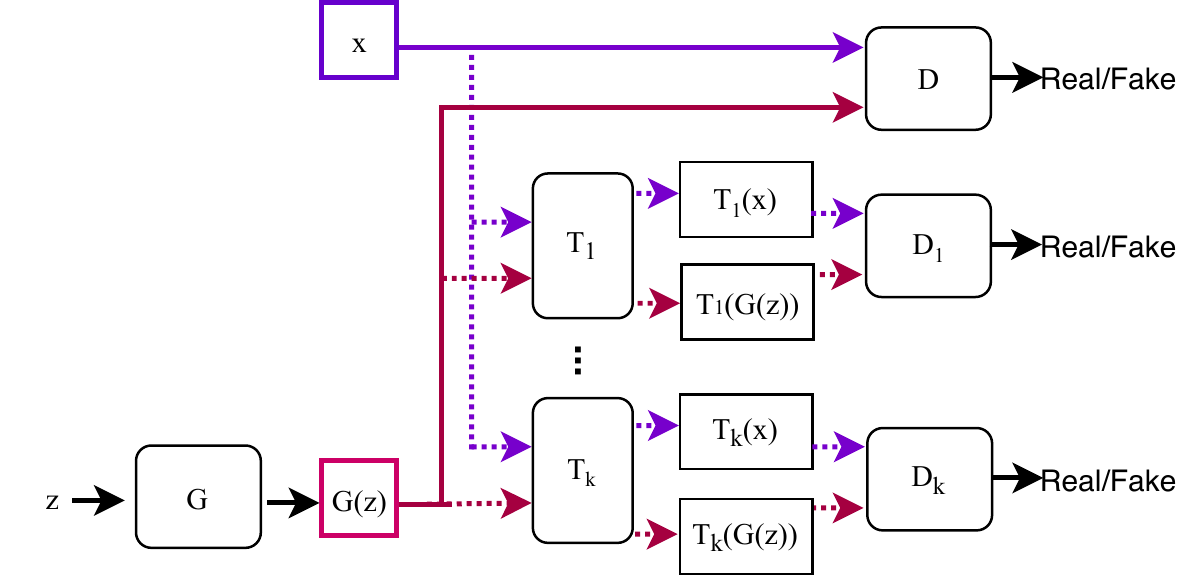}
    \caption{The modified models with K branches from DAG: k = 2,...,K without data augmentation in generator learning (represented by dot lines -- note that these are used in training $D_k$).}
    \label{fig_no_g_dag}
\end{figure}

First, we compare our DAG to IDA (see Table \ref{distgan_on_shared_weights}). The results suggest that IDA is not as good as  DAG, which means that when JS divergence is not preserved (i.e., minimizing lower-bounds in the case of IDA), the performance is degraded. Second,
when training the {\em generator}, 
we remove branches $T_k$, i.e. in generator training, no augmented sample is used (Fig. \ref{fig_no_g_dag}).
We use DAG models with rotation as the baselines and others are kept exactly the same as DAG. Substantial degradation occurs as shown in Table \ref{gan_without_g_dag}. This confirms the significance of augmented samples in \textit{generator learning.} %Refer to Appendix \ref{appendix_c} for studies of  data augmentation (Appendix \ref{importance_of_da_in_dag}) and the impact of the number of branches K (Appendix \ref{dag_k_study}).

\begin{table*}
  \caption{FID of DistGAN + DAG (rotation) and SS-GAN + DAG (rotation) with and without augmented samples in generator learning. ``-G": no augmented samples in G learning.}
  \label{ablation_study_md_distgan}
  \centering
  \begin{tabular}{l|c|c|c|c}
    \toprule
    \textbf{Methods} & DistGAN+DAG & DistGAN+DAG (-G) & SSGAN+DAG & SSGAN+DAG (-G)\\
    \hline
    \textbf{FID} & 23.7 & 30.1 & 25.2 & 31.5\\
    \bottomrule
  \end{tabular}
  \label{gan_without_g_dag}
\end{table*}

%\subsection{The importance of transformations in generator learning of DAG}
%\label{dag_no_g}
%
%To validate the importance of transformation in generator learning of DAG, we remove $T_k$ when training DAG for the generator as shown in Fig. \ref{fig_no_g_dag}. Note that we still train discriminator exactly as our DAG.

\subsubsection{The importance of data augmentation in our DAG}
\label{importance_of_da_in_dag}

Tables \ref{distgan_md} and \ref{ssgan_md} represent the additional results of other DAG methods comparing to the multiple discriminators (MD) variant.
MD is exactly the same as DAG as in Fig. \ref{dag_models}b, except that all the transformations T$_k$ are removed, i.e. MD does not apply augmented data.
We use K = 4 branches for all DAG methods. The experiments are with two baseline models: DistGAN and  SSGAN. We train DistGAN + MD and SS-GAN + MD on full (100\%) CIFAR-10 dataset. Using MD indeed slightly improves the performance of Baseline, but the performance is substantially improved further as adding any augmentation technique (DAG). {\em This study  verifies the importance of augmentation techniques and our DAG in the improvement of GAN baseline models.} We use small DCGAN (Appendix \ref{appendix_d}) for this experiment.\\

\begin{table}
  \footnotesize
  \caption{FID of DistGAN + MD compared with DistGAN Baseline and our DistGAN + DAG methods.}
  \label{ablation_study_md_distgan}
  \centering
  \begin{tabular}{ll}
    \toprule
    \textbf{Methods} & \textbf{FID} \\
    \midrule
    DistGAN  & 29.6 \\
    \hline
    {DistGAN + MD}  & {27.8} \\
    \hline
    DistGAN + DAG (rotation)  & 23.7 \\
    \hline
    DistGAN + DAG (flipping)  & 25.0 \\
    \hline
    DistGAN + DAG (cropping)  & 24.2 \\
    \hline
    DistGAN + DAG (translation)  & 25.5 \\
    \hline
    DistGAN + DAG (flipping+rotation)  & 23.3 \\
    \bottomrule
  \end{tabular}
  \label{distgan_md}
\end{table}

\begin{table}
  \footnotesize
  \caption{FID of SSGAN + MD compared with SSGAN Baseline and our SSGAN + DAG methods.}
  \label{ablation_study_md_ssgan}
  \centering
  \begin{tabular}{ll}
    \toprule
    \textbf{Methods} & \textbf{FID} \\
    \midrule
    SSGAN       & 28.0 \\
    \hline
    {SSGAN + MD}  & {27.2} \\
    \hline
    SSGAN + DAG (rotation)  & 25.2 \\
    \hline
    SSGAN + DAG (flipping)  & 25.9 \\
    \hline
    SSGAN + DAG (cropping)  & 23.9 \\
    \hline
    SSGAN + DAG (translation)  & 26.3 \\
    \hline
    SSGAN + DAG (flipping+rotation)  & 25.2 \\    
    \bottomrule
  \end{tabular}
  \label{ssgan_md}
\end{table}

\subsubsection{The ablation study on the number of branches K of DAG}
\label{dag_k_study}

We conduct the ablation study on the number of branches K in our DAG, we note that using large K is adhesive to combine more augmentations since each augmentation has the limit number of invertible transformations in practice, i.e. 4 for rotations (Table \ref{da_techniques}). The Dist-GAN + DAG model is used for this study. In general, we observe that the larger K is (by simply combining with other augmentations on top of the current ones), the better FID scores DAG gets as shown in Table \ref{distgan_on_k}. However, there is a trade-off between the accuracy and processing time as increasing the number of branches K. (Refer to more details about the training time in Section \ref{implementation_computation}). We use small DCGAN (Appendix \ref{appendix_d}) for this experiment.

\begin{table}
  \footnotesize
  \caption{The ablation study on the number of branches K in our DAG model. We use Dist-GAN + DAG as the baseline for this study.}
  \label{ablation_study_md_distgan}
  \centering
  \begin{tabular}{ll}
    \toprule
    \textbf{Number of branches} & \textbf{FID} \\
    \midrule
    K = 4 (1 identity + 3 rotations) & 23.7 \\
    \hline
    %K = 7 (1 identity + 3 rotations + 3 flippings) & 24.1 \\
    %\hline
    K = 7 (1 identity + 3 rotations + 3 flippings) & 23.1 \\
    \hline
    %K = 7 (1 identity + 3 rotations + 3 croppings) & ... \\     
    %\hline
    K = 10 (1 identity + 3 rotations + 3 flippings + 3 croppings) & 22.4 \\ 
    %\hline
    %K = 13 (1 identity + 3 rotations + 3 flippings + 3 croppings + 3 translations) & ... \\     
    \bottomrule
  \end{tabular}
  \label{distgan_on_k}
\end{table}

\subsection{Data Augmentation optimized for GAN}
\label{exp_data_augmentation}

\begin{figure*}
  \centering
  \includegraphics[width=3.5cm,keepaspectratio]{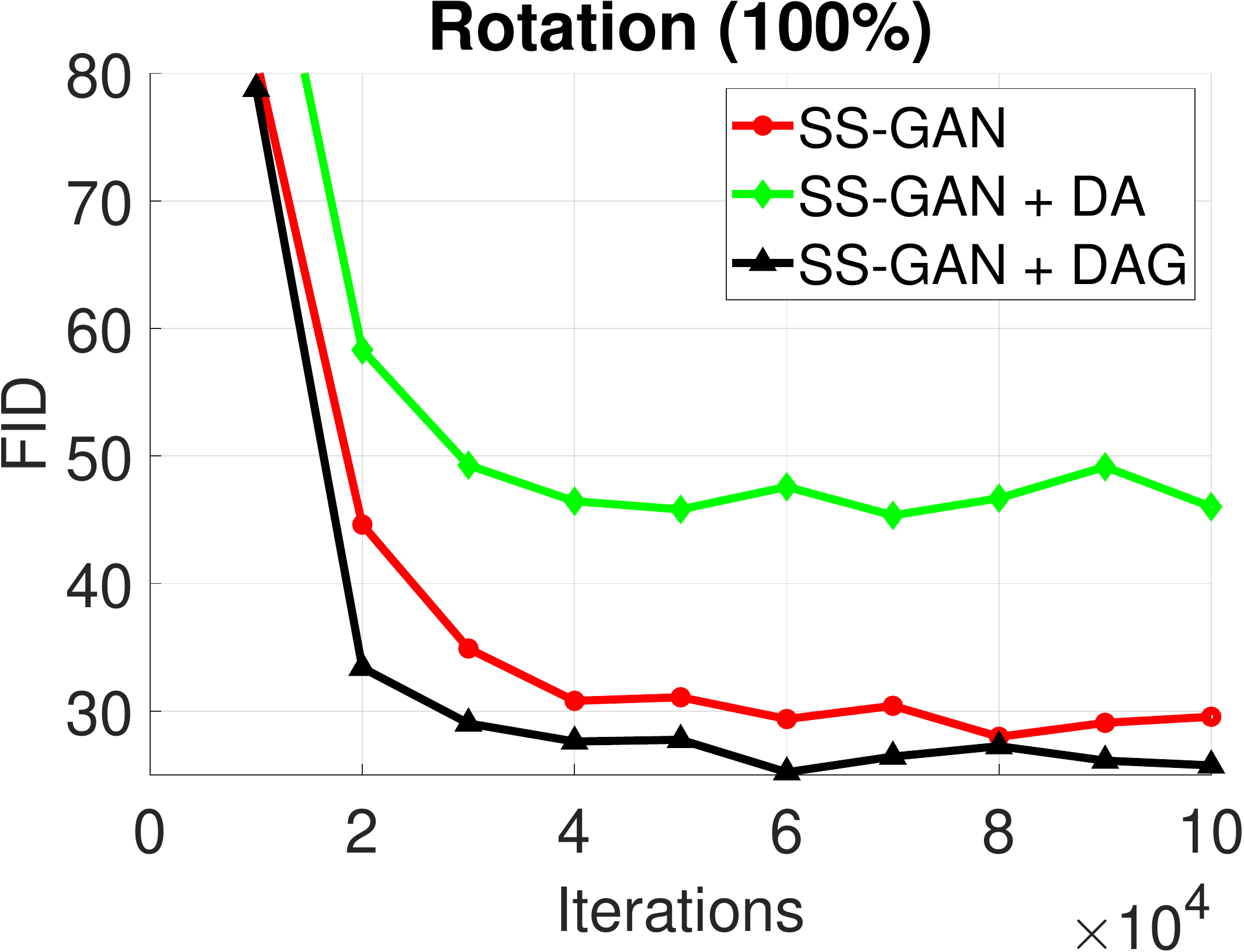}
  \includegraphics[width=3.5cm,keepaspectratio]{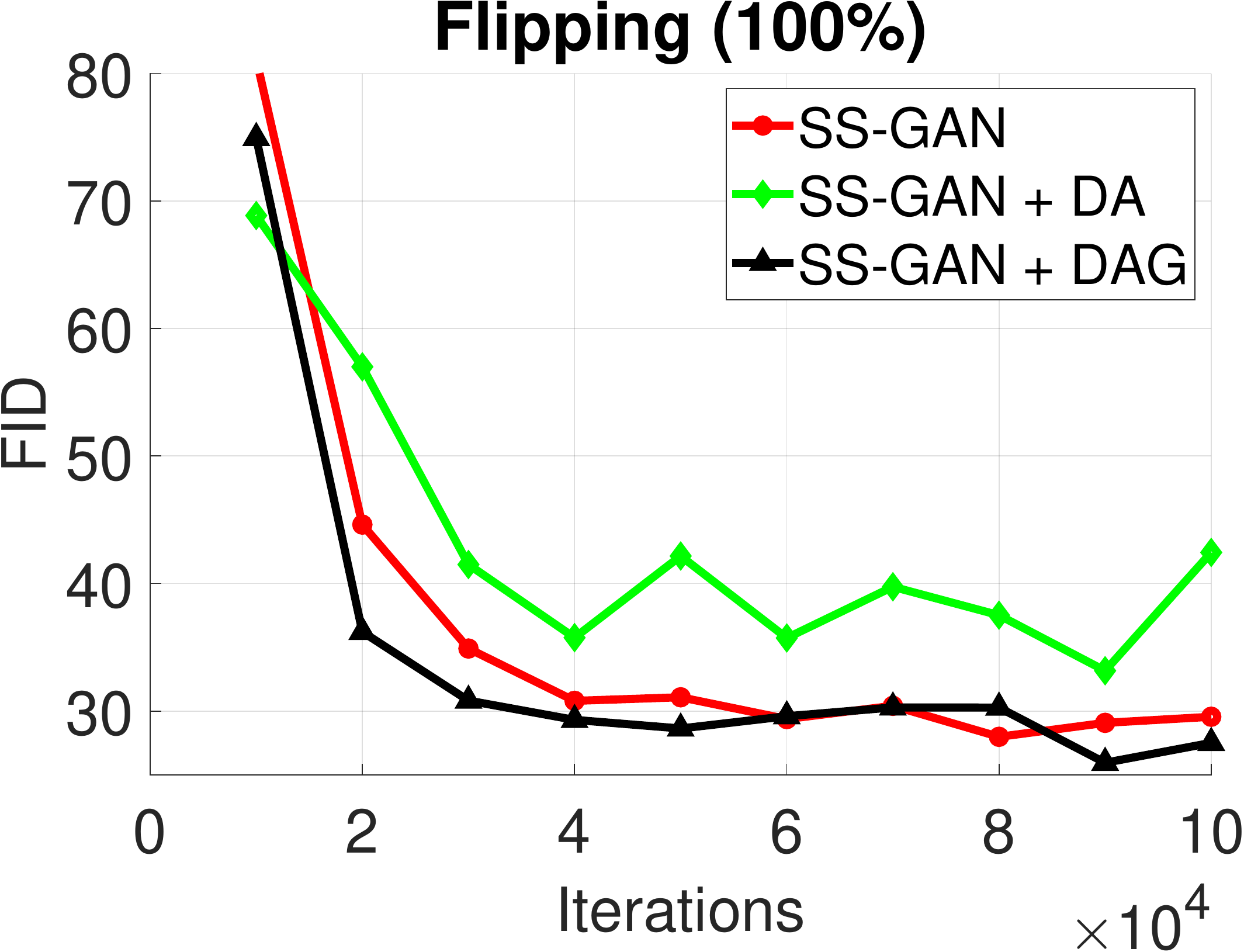}
  \includegraphics[width=3.5cm,keepaspectratio]{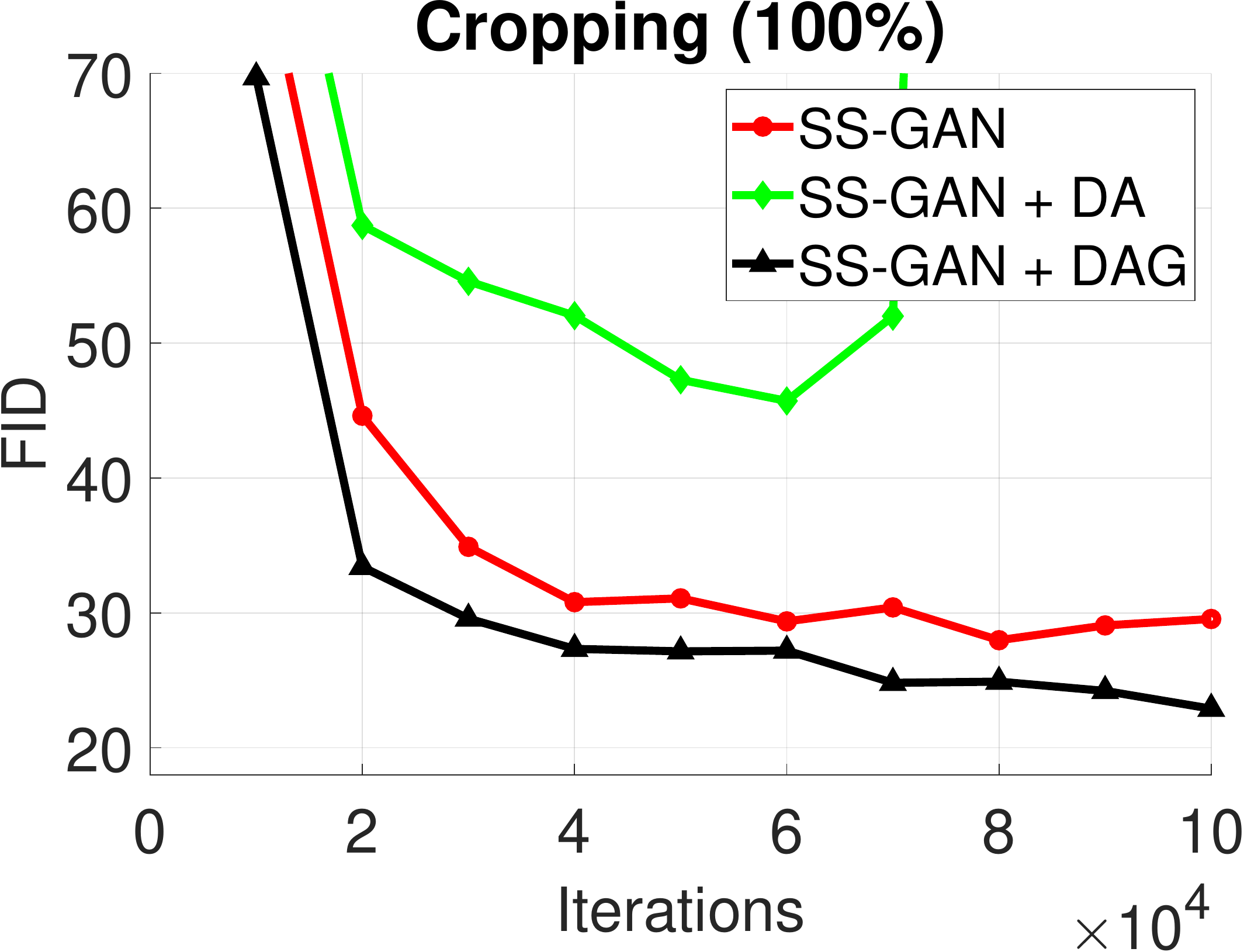}
  \includegraphics[width=3.5cm,keepaspectratio]{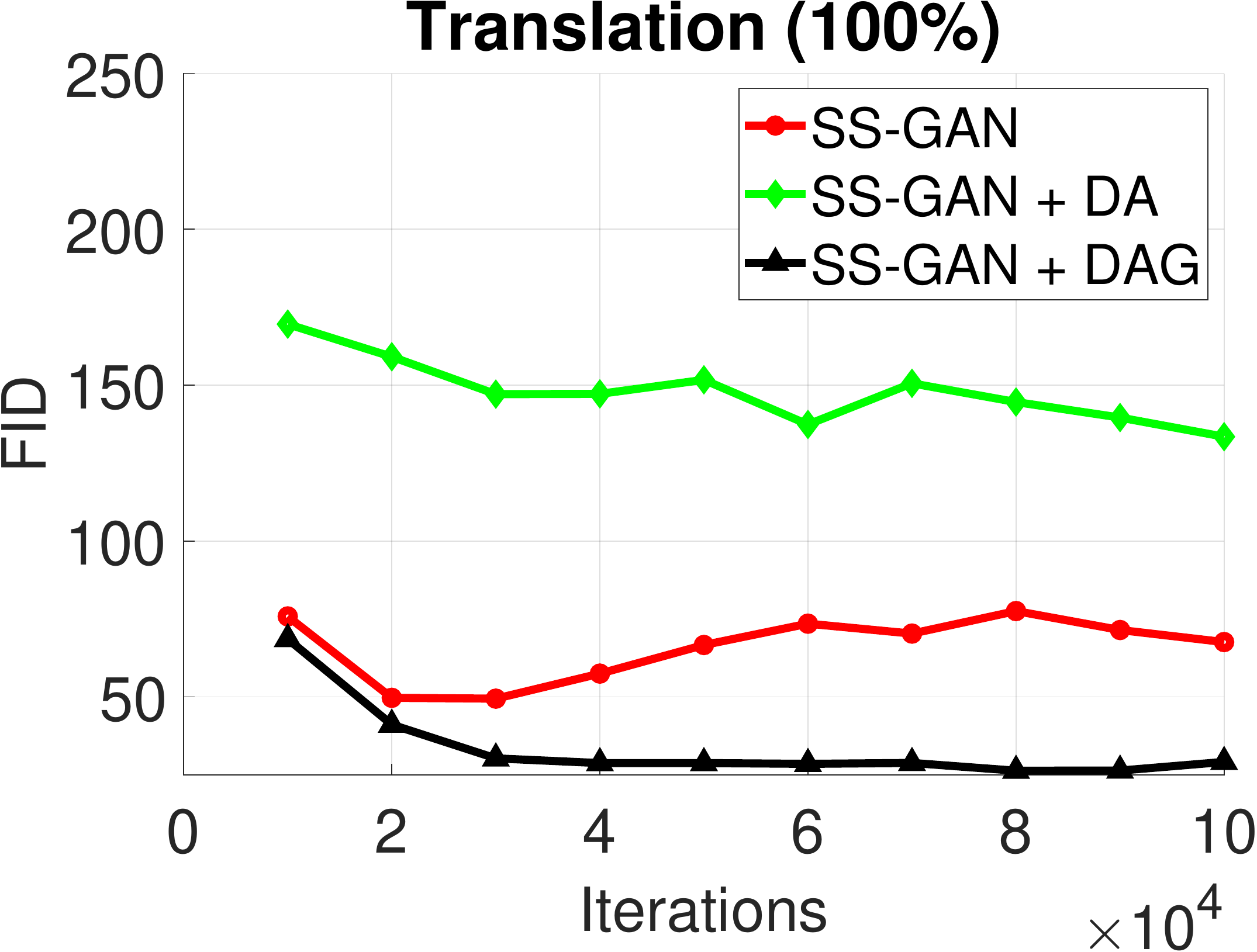}
  \includegraphics[width=3.5cm,keepaspectratio]{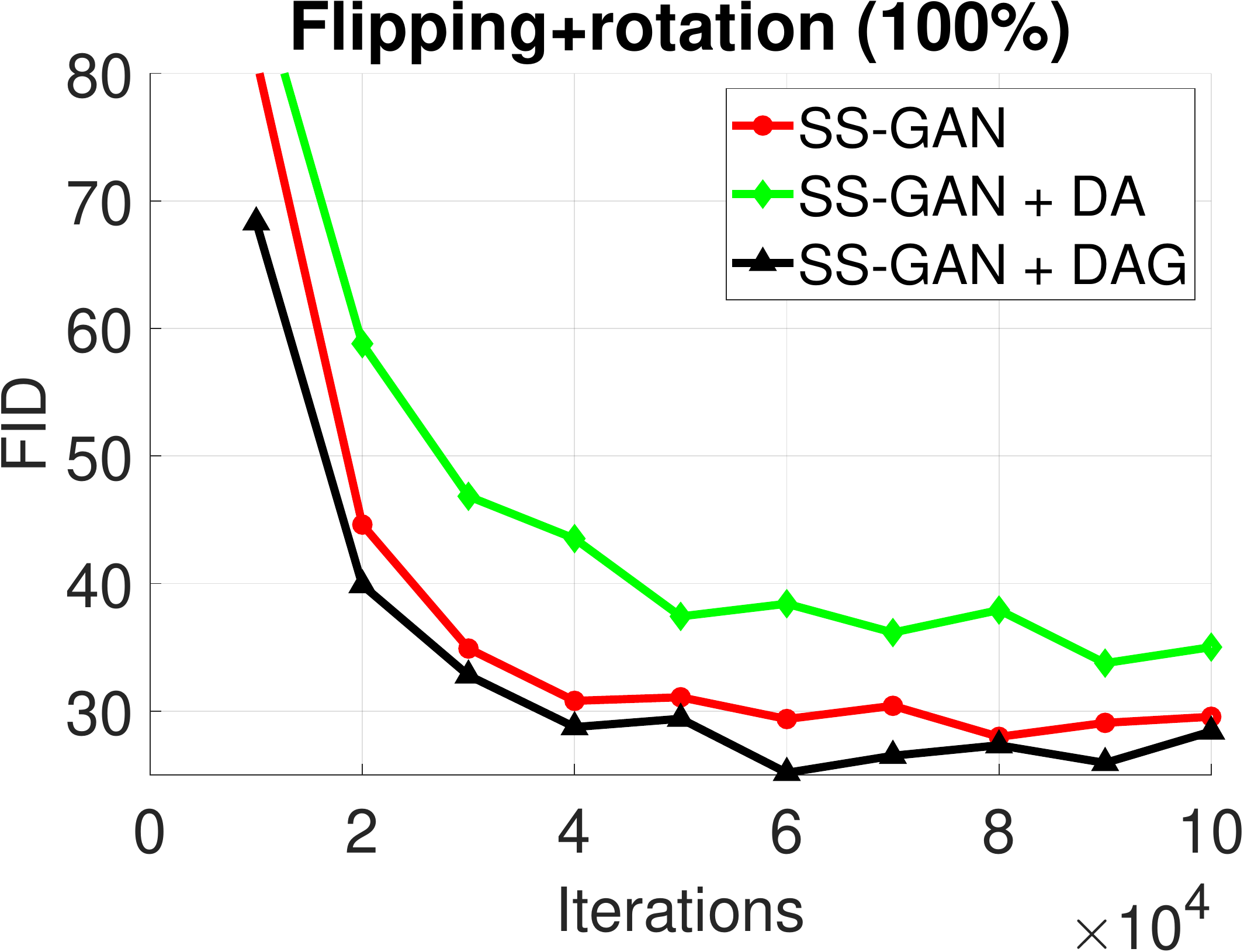}\\
  \includegraphics[width=3.5cm,keepaspectratio]{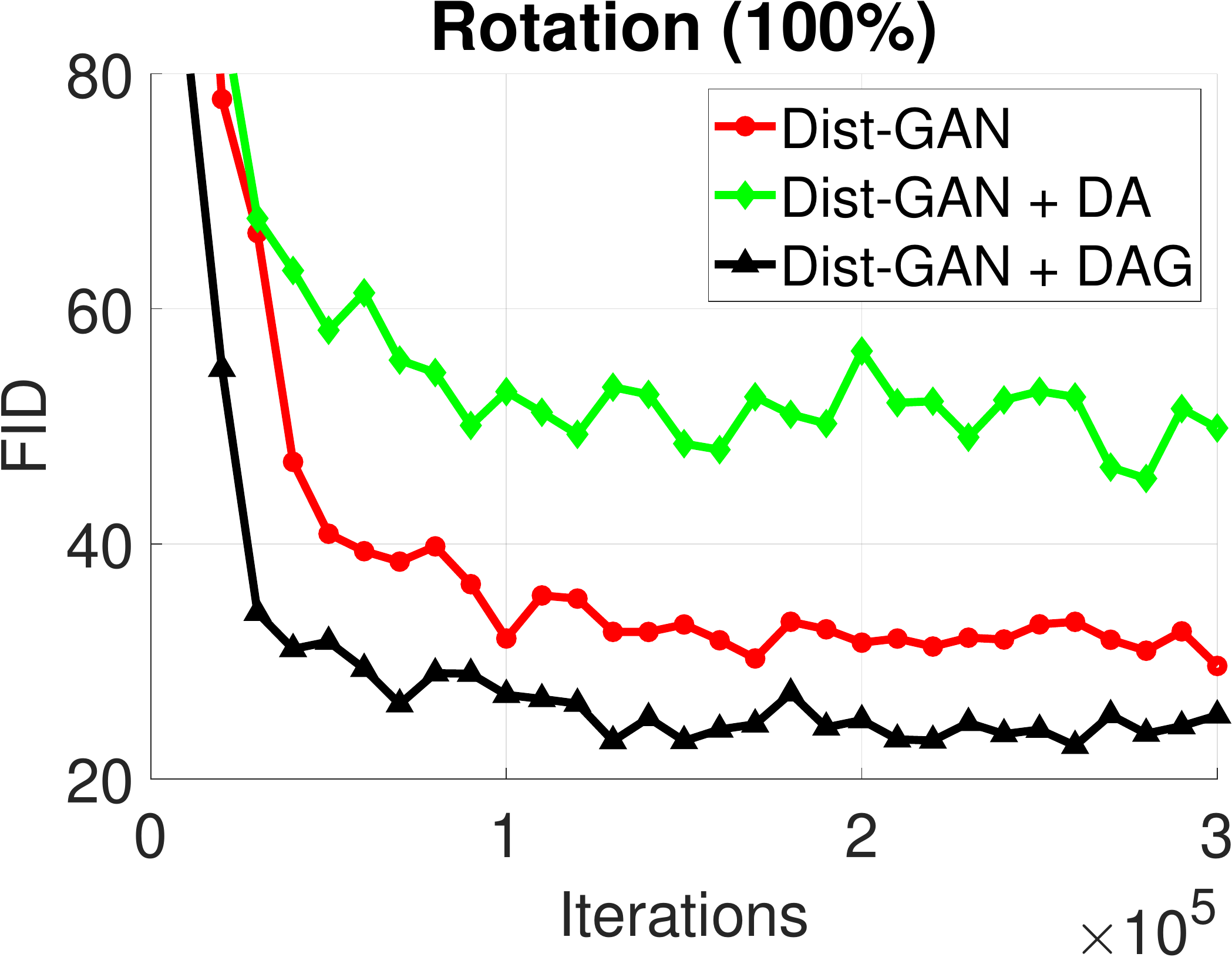}
  \includegraphics[width=3.5cm,keepaspectratio]{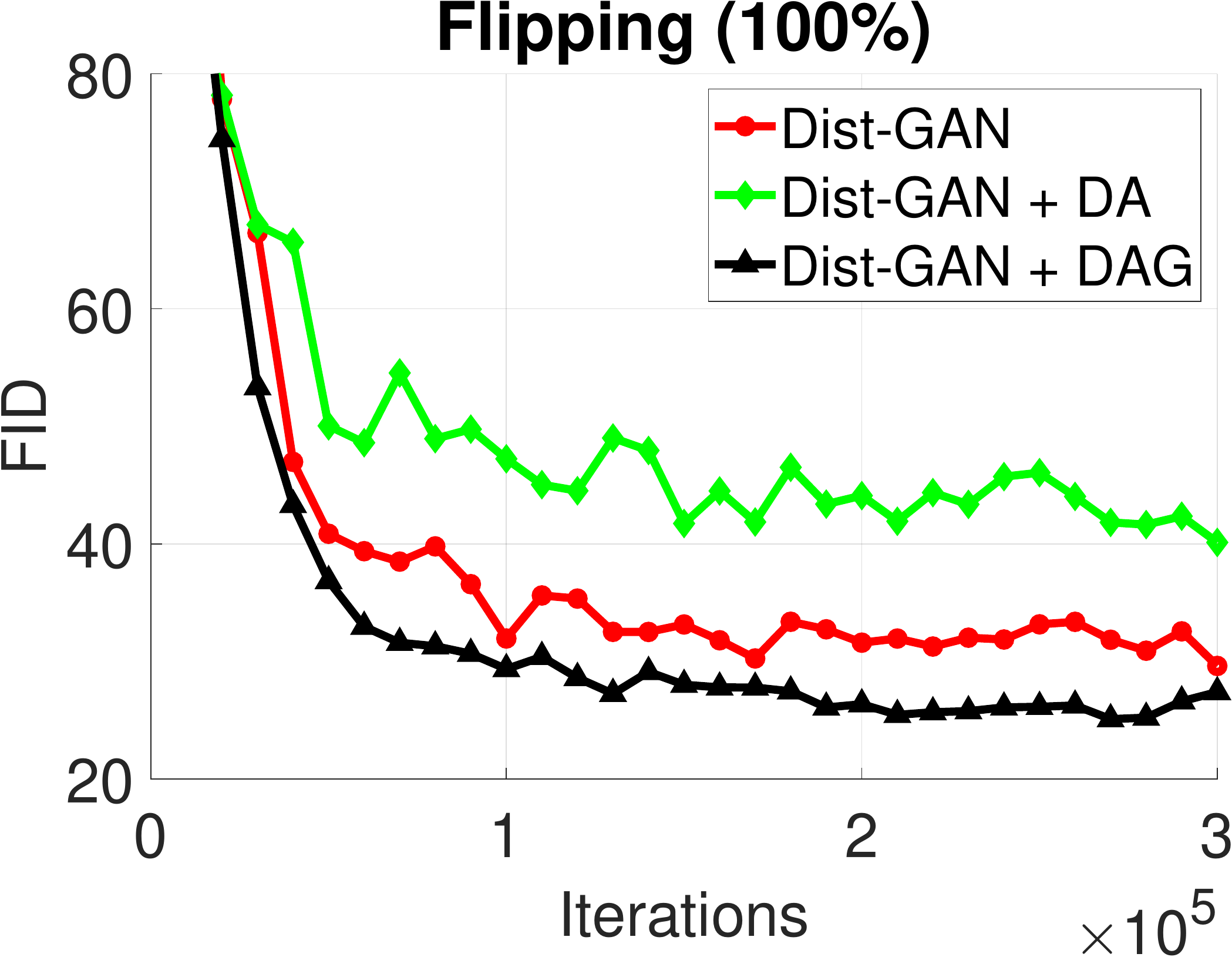} 
  \includegraphics[width=3.5cm,keepaspectratio]{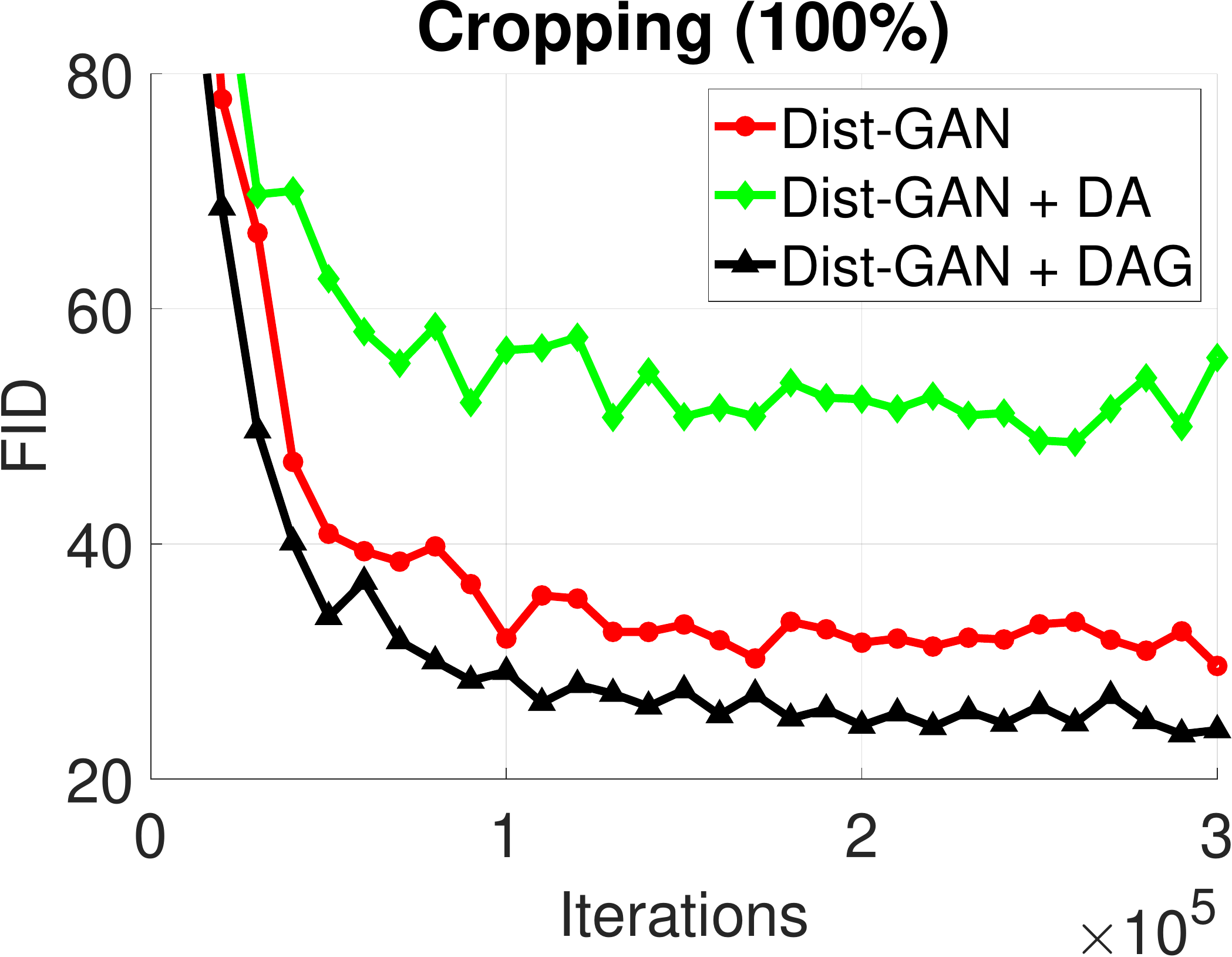}
  \includegraphics[width=3.5cm,keepaspectratio]{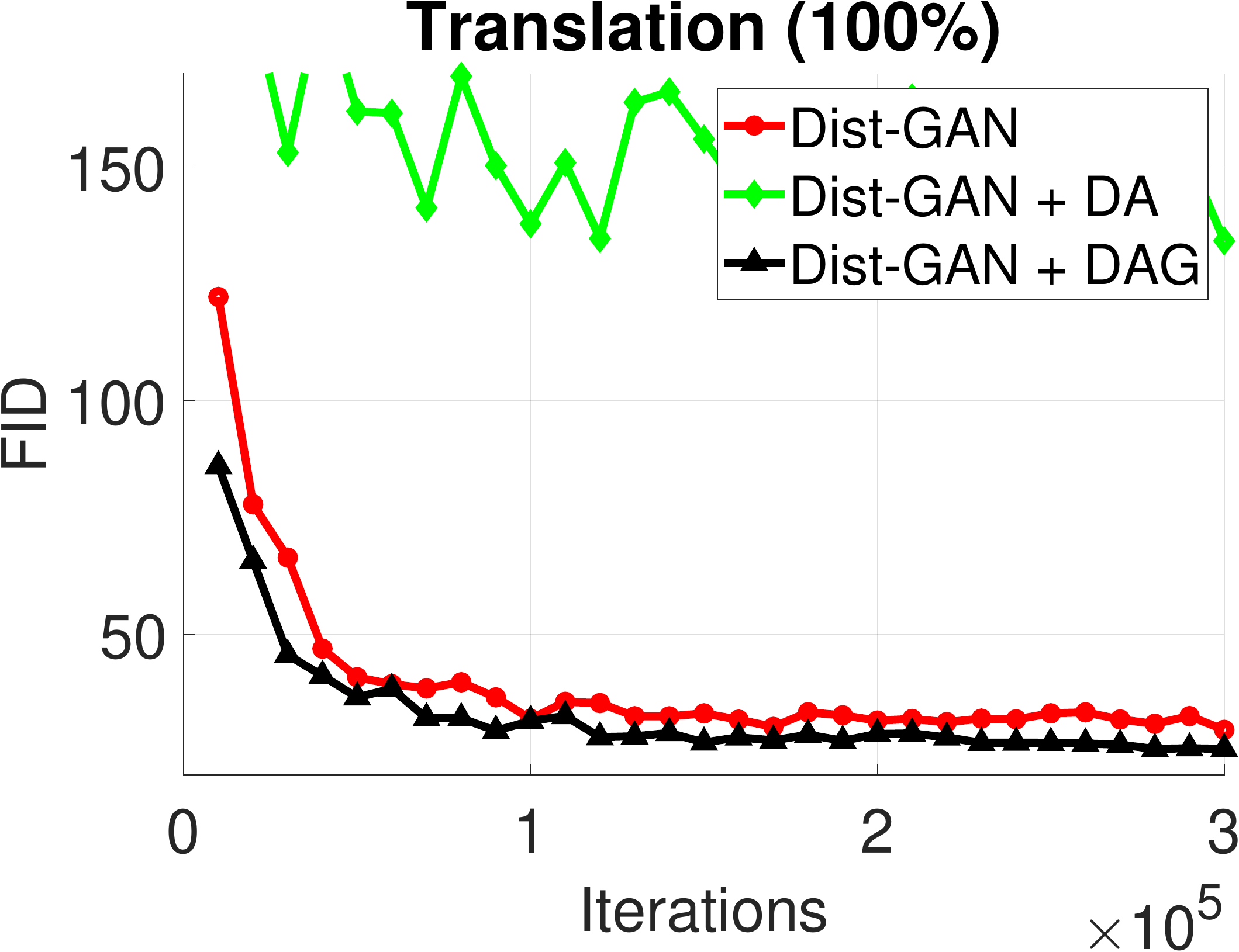} 
  \includegraphics[width=3.5cm,keepaspectratio]{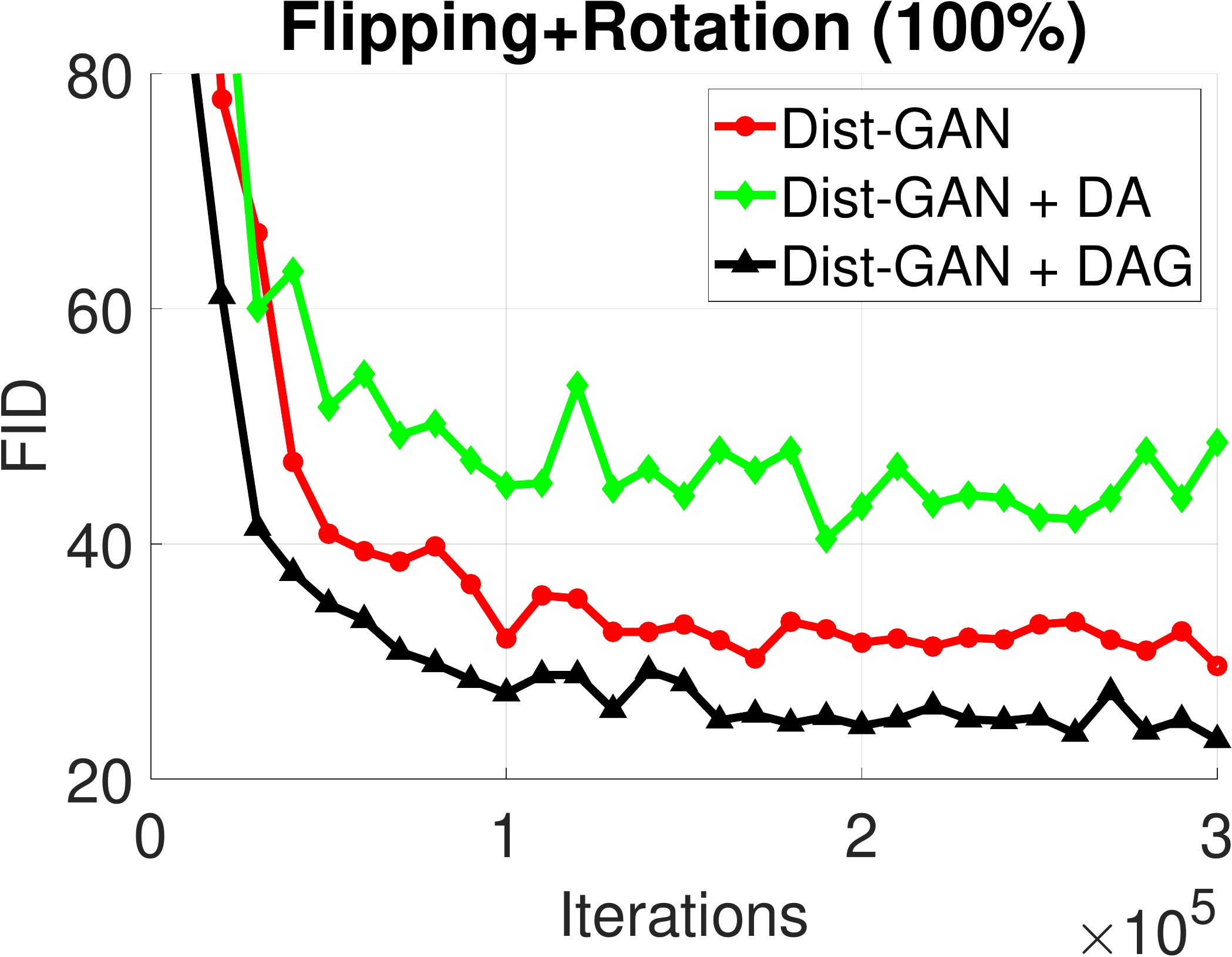}

  \caption{Comparing DA and our proposed DAG with SS-GAN \cite{chen-arxiv-2018} (first row) and Dist-GAN \cite{tran-eccv-2018} (second row) baselines on full dataset (100\%). Left to right columns: rotation, flipping, cropping, translation, and flipping+rotation. The horizontal axis is the number of training iterations, and the vertical axis is the FID score.}
  \label{data_argumentation_100}
\end{figure*}

\begin{figure*}
  \centering
  \includegraphics[width=3.5cm,keepaspectratio]{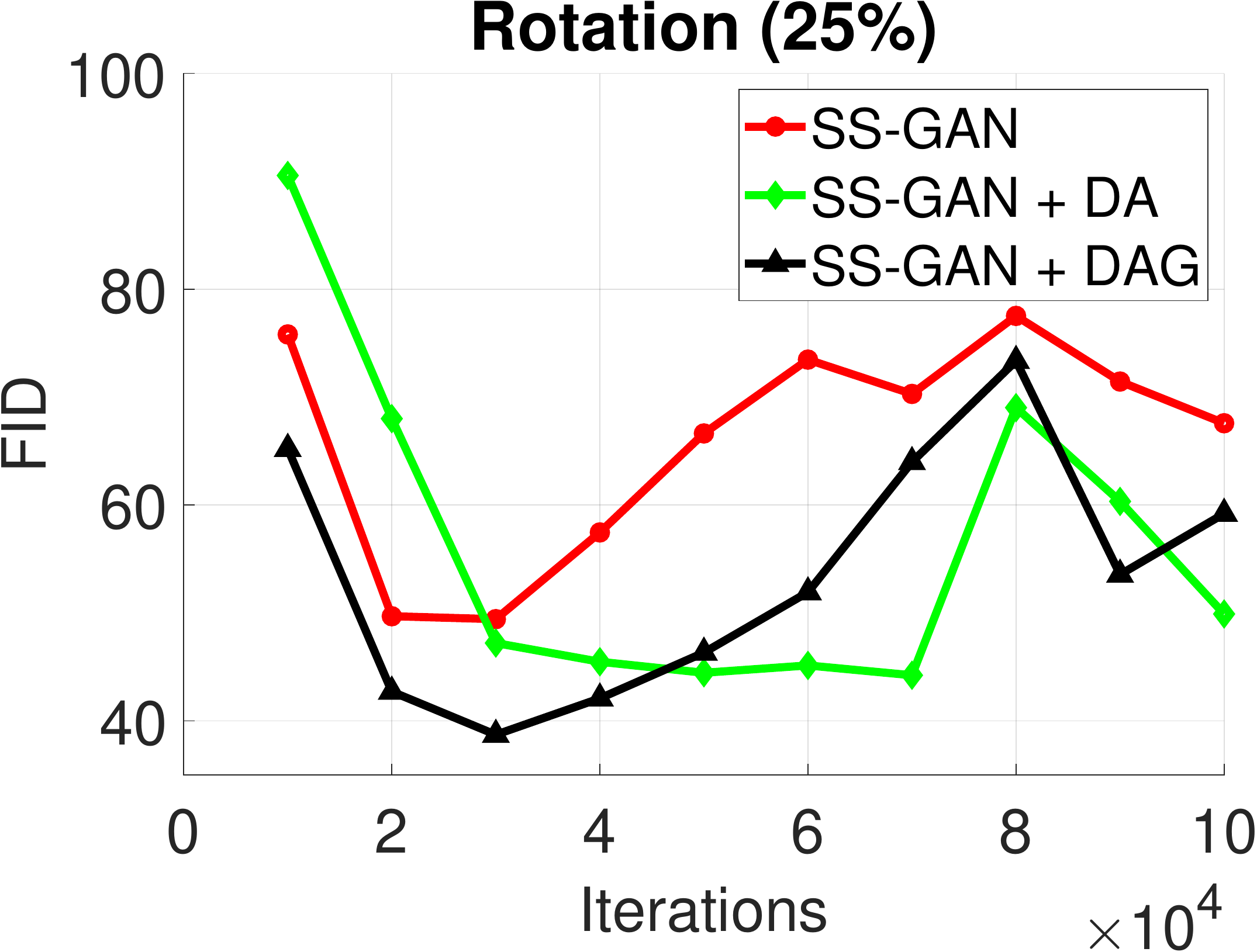}
  \includegraphics[width=3.5cm,keepaspectratio]{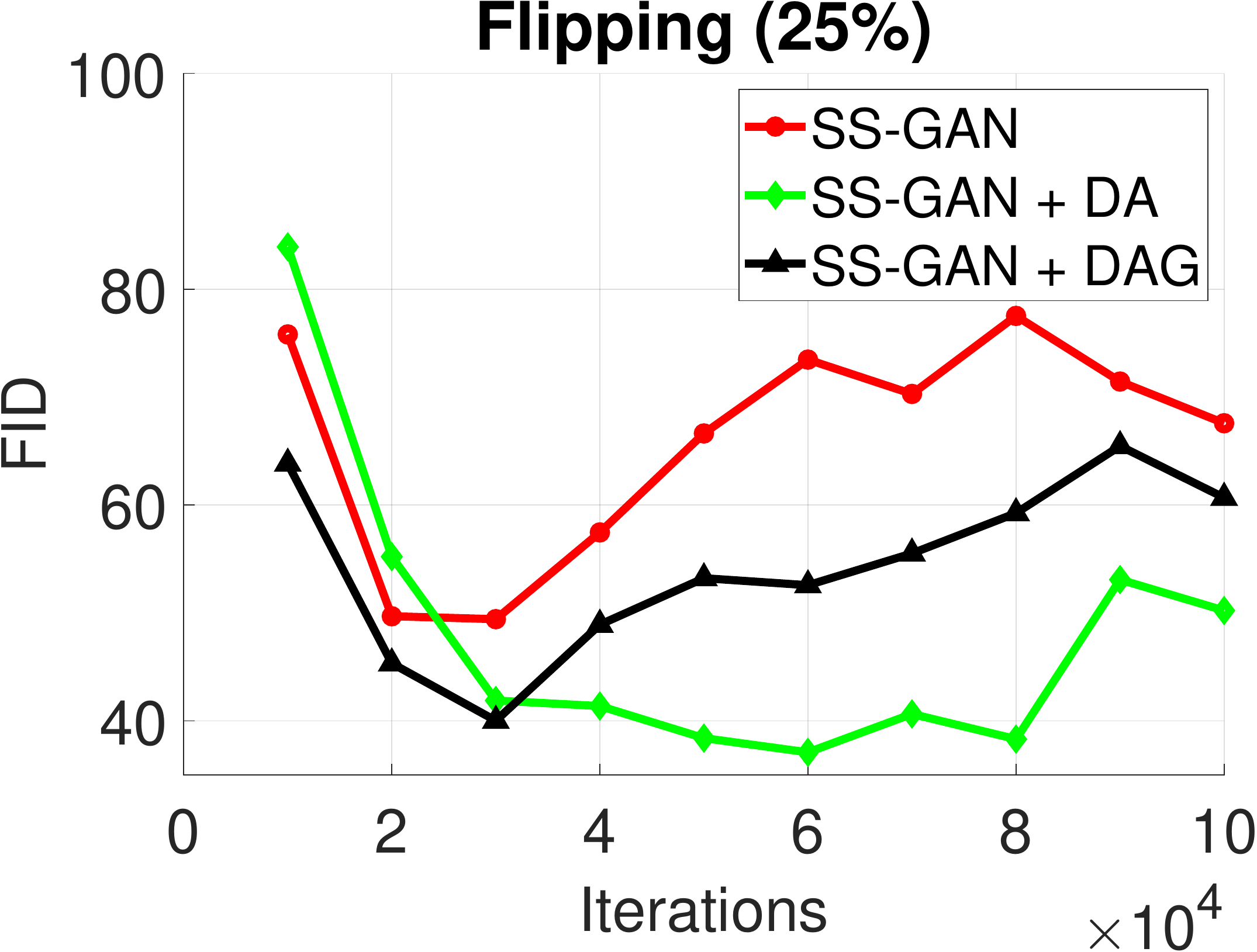}
  \includegraphics[width=3.5cm,keepaspectratio]{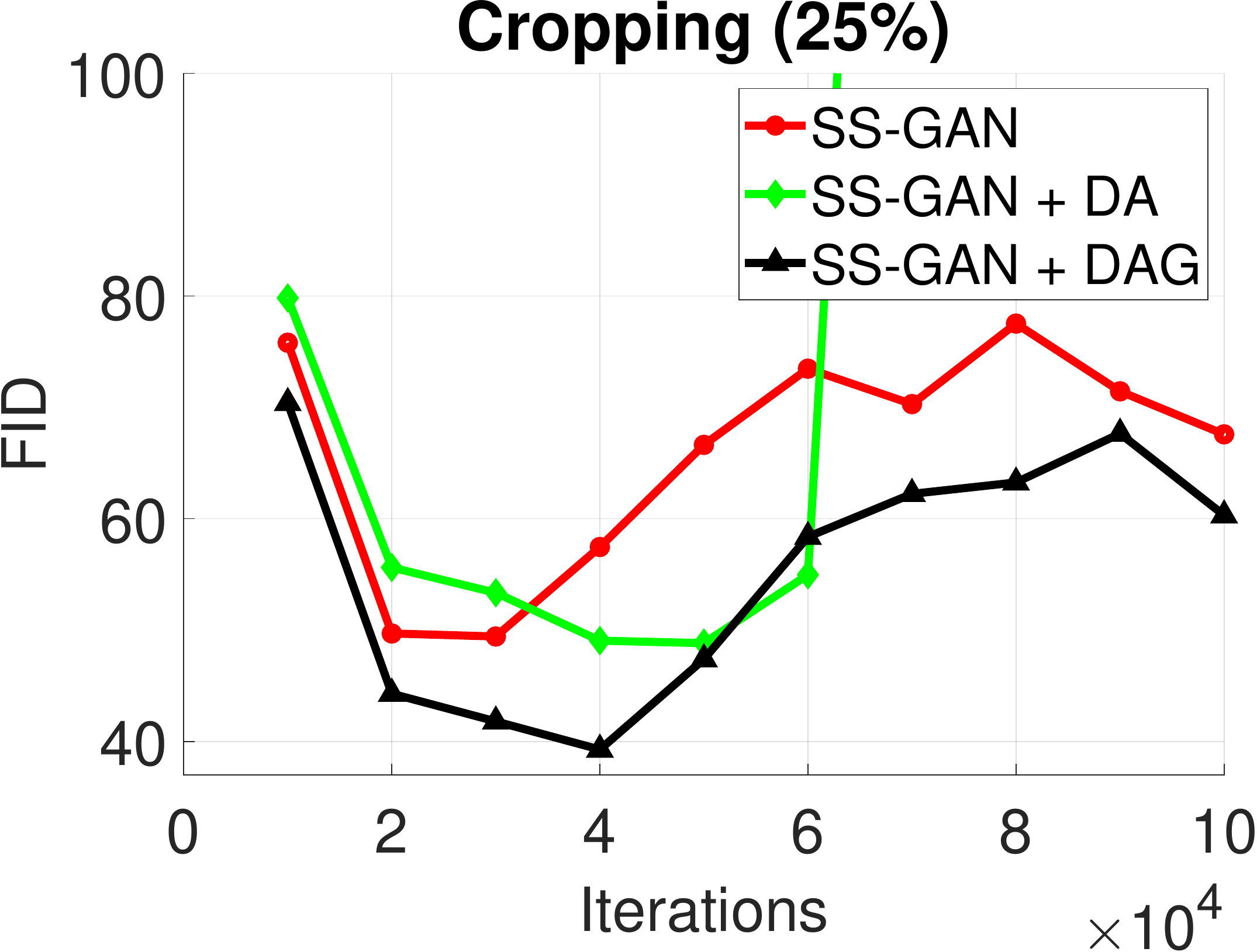}
  \includegraphics[width=3.5cm,keepaspectratio]{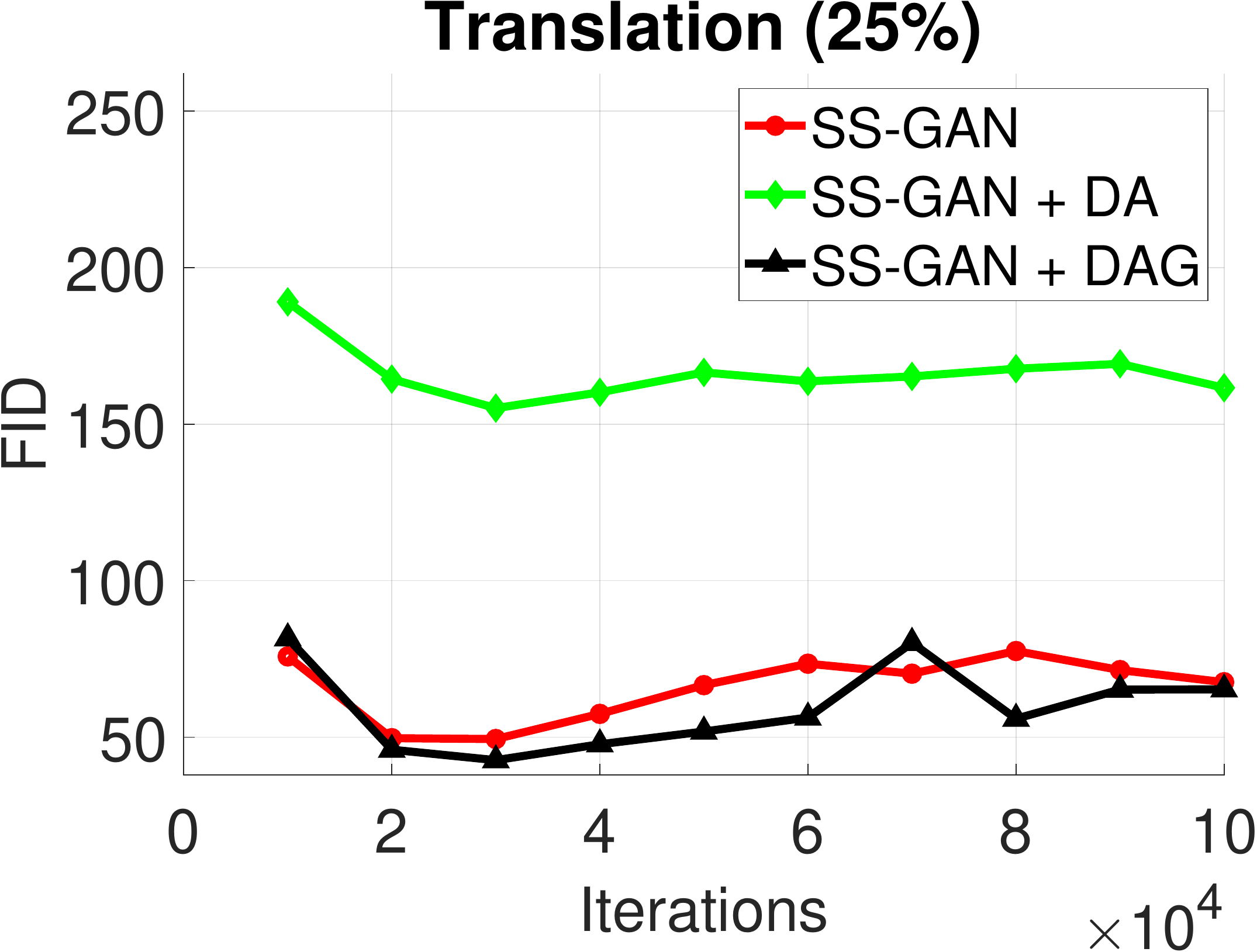}
  \includegraphics[width=3.5cm,keepaspectratio]{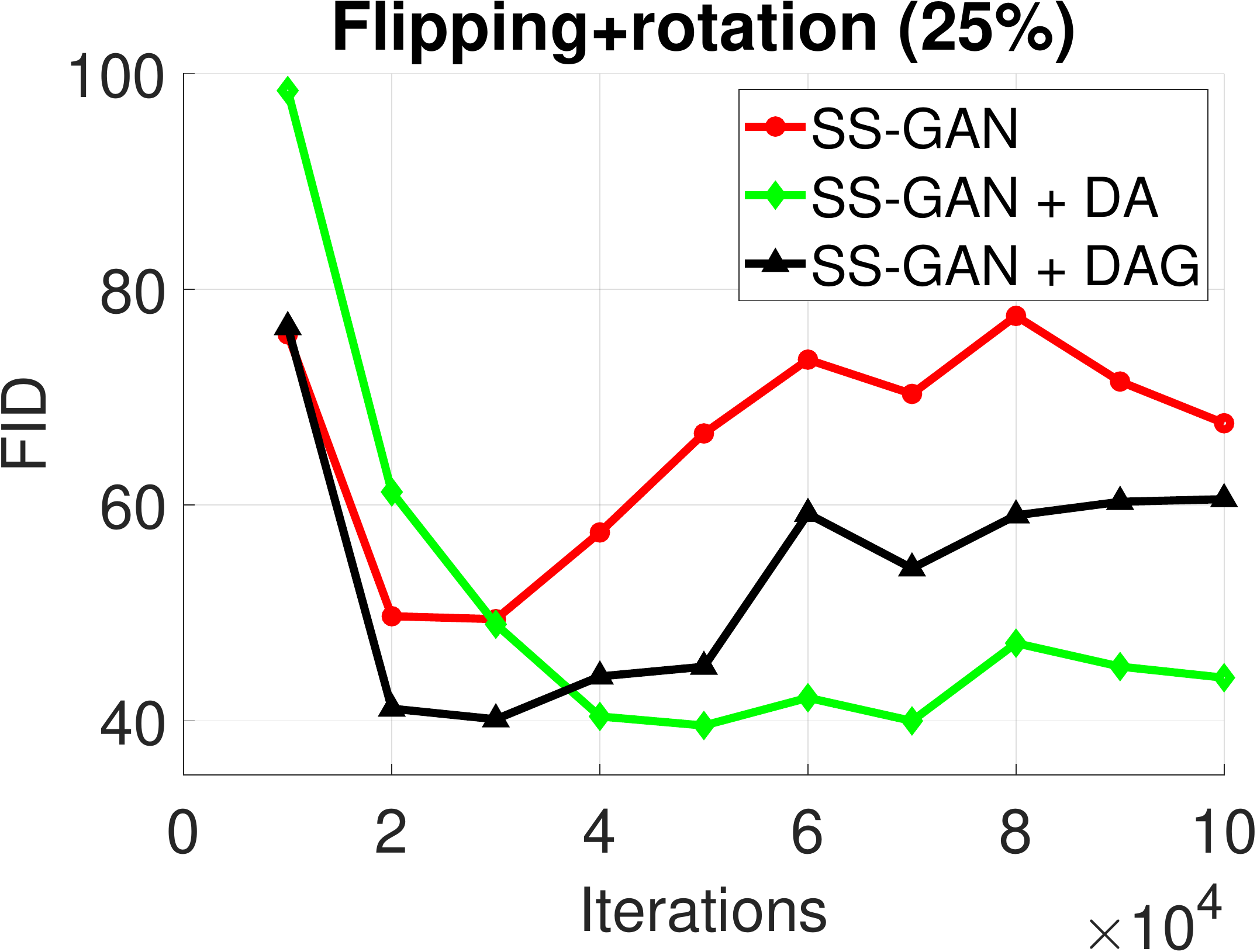}\\
  \includegraphics[width=3.5cm,keepaspectratio]{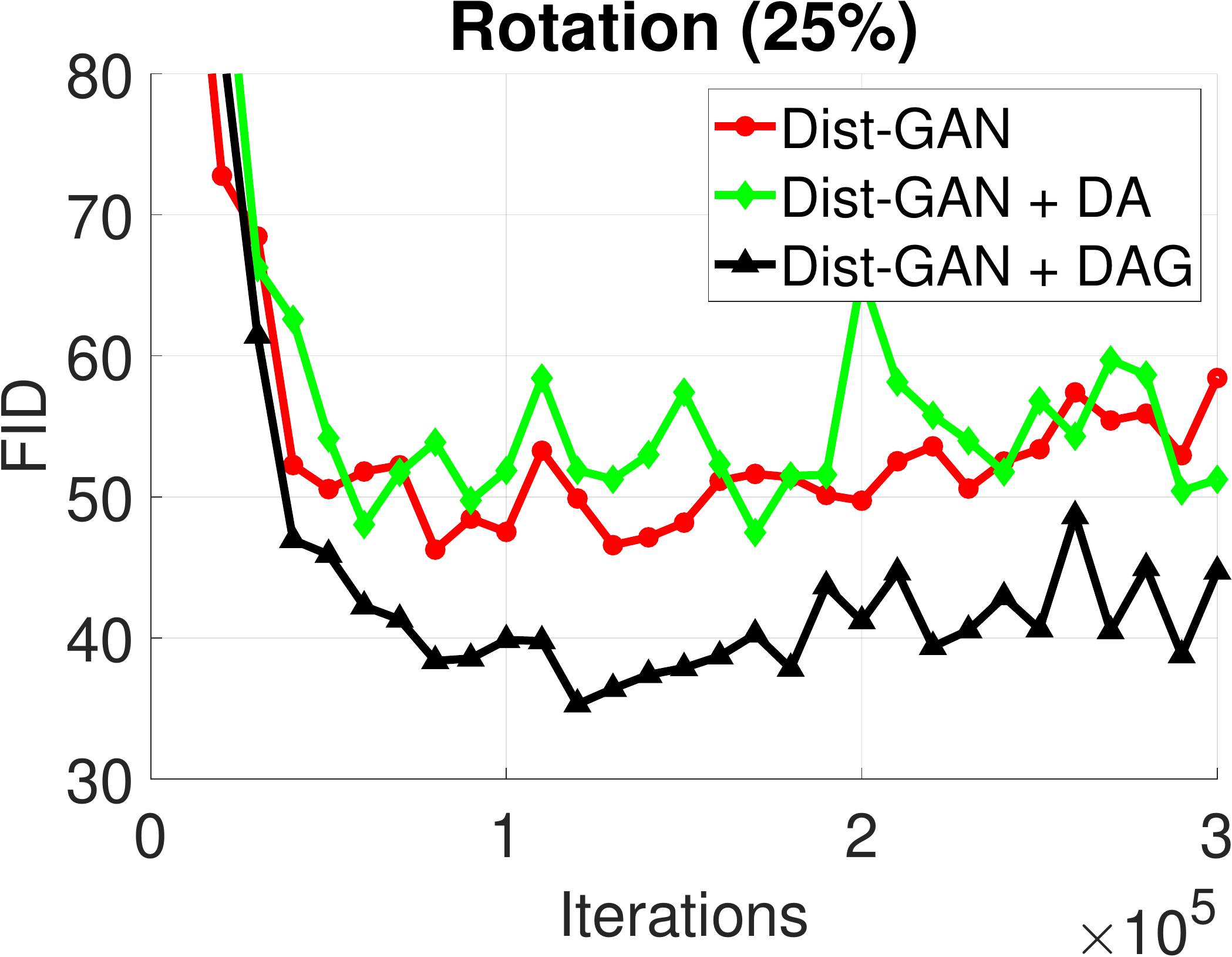}
  \includegraphics[width=3.5cm,keepaspectratio]{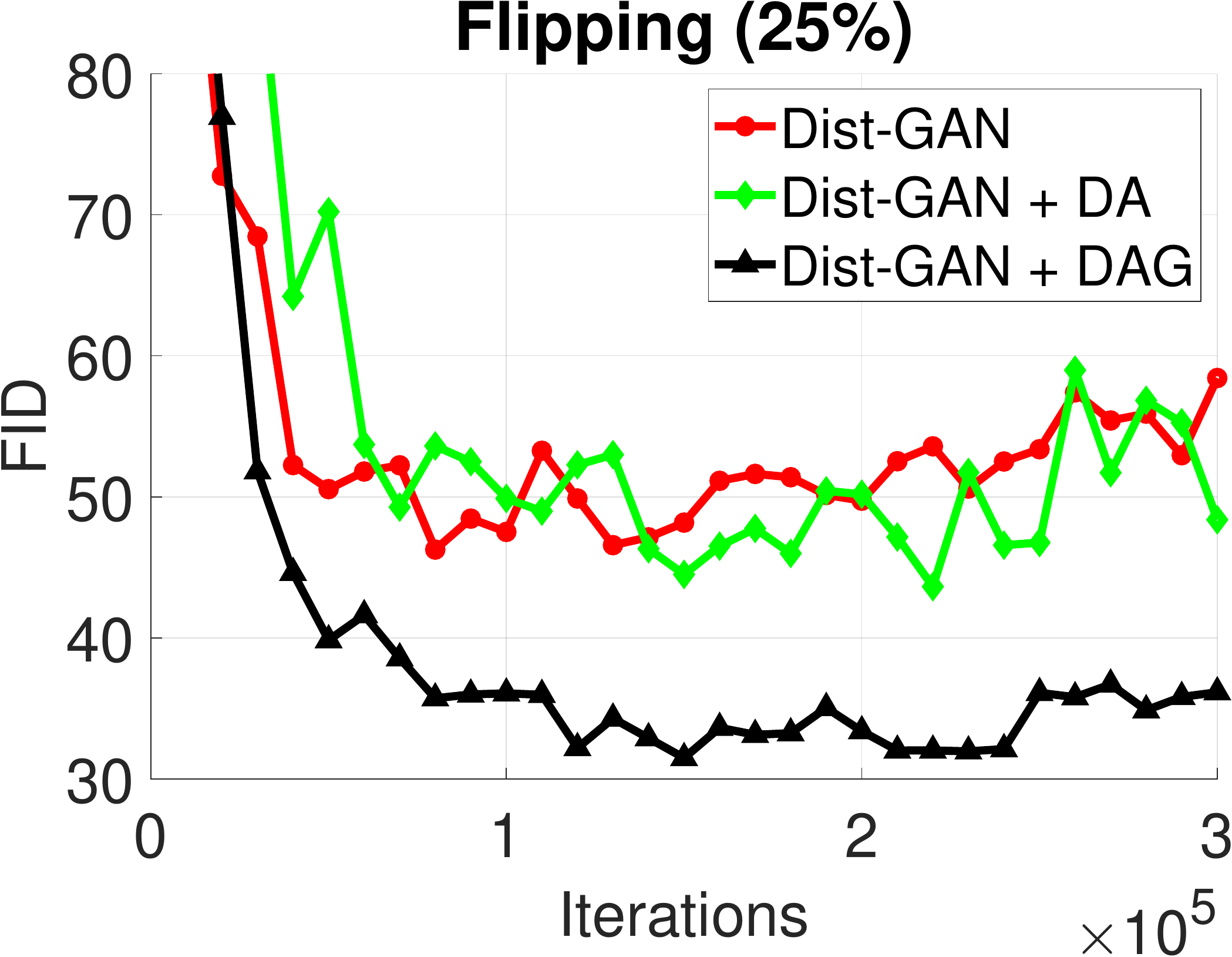} 
  \includegraphics[width=3.5cm,keepaspectratio]{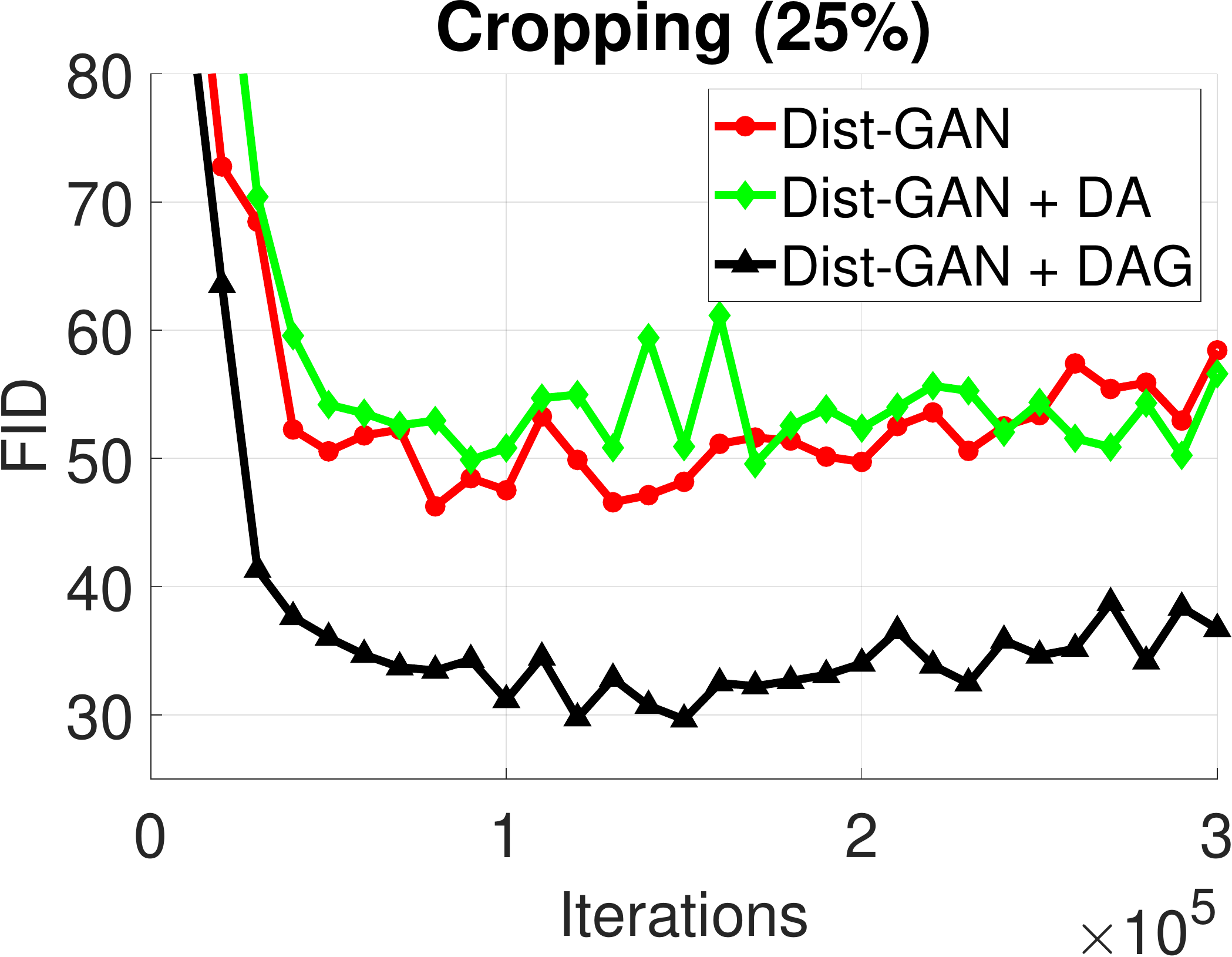} \includegraphics[width=3.5cm,keepaspectratio]{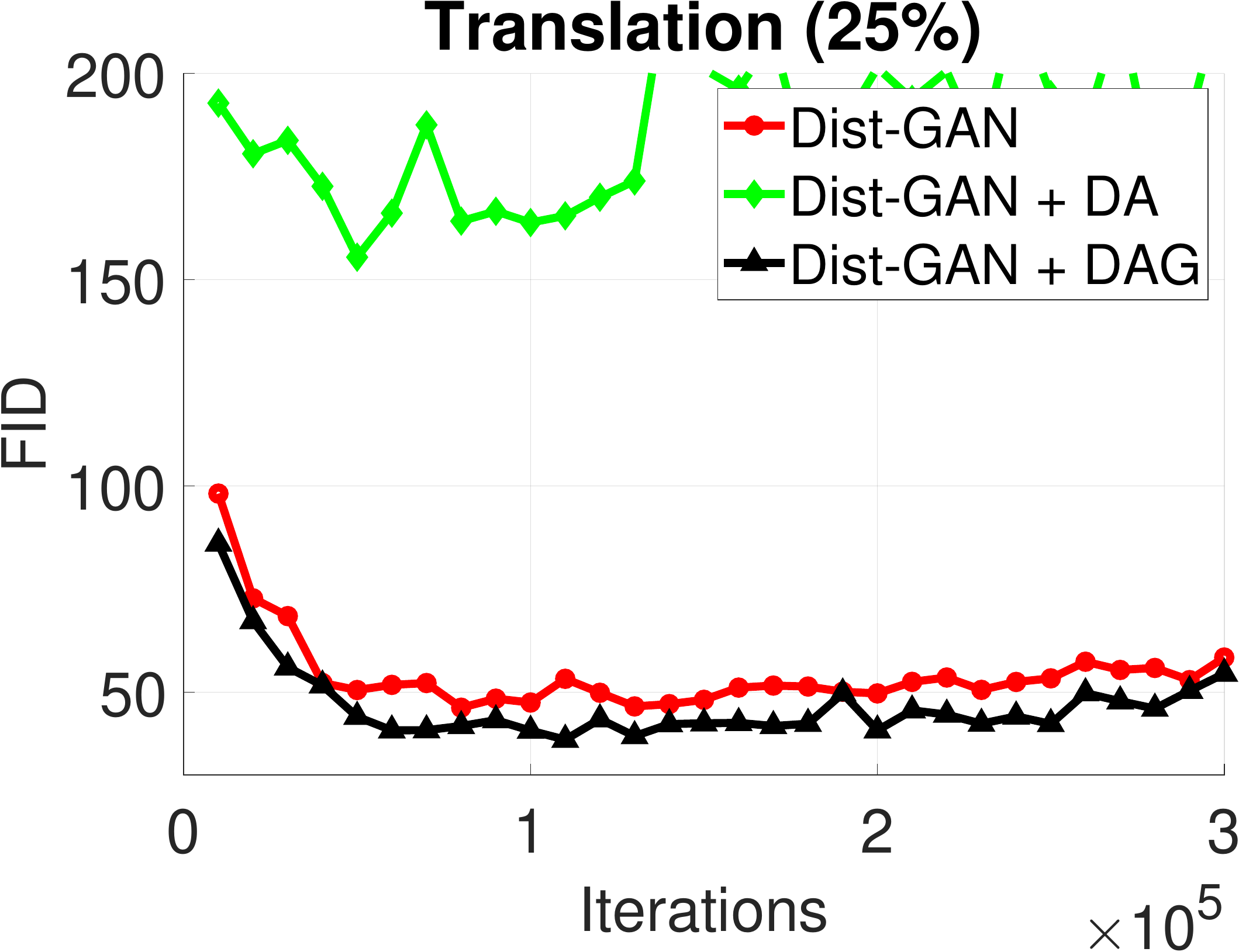}
  \includegraphics[width=3.5cm,keepaspectratio]{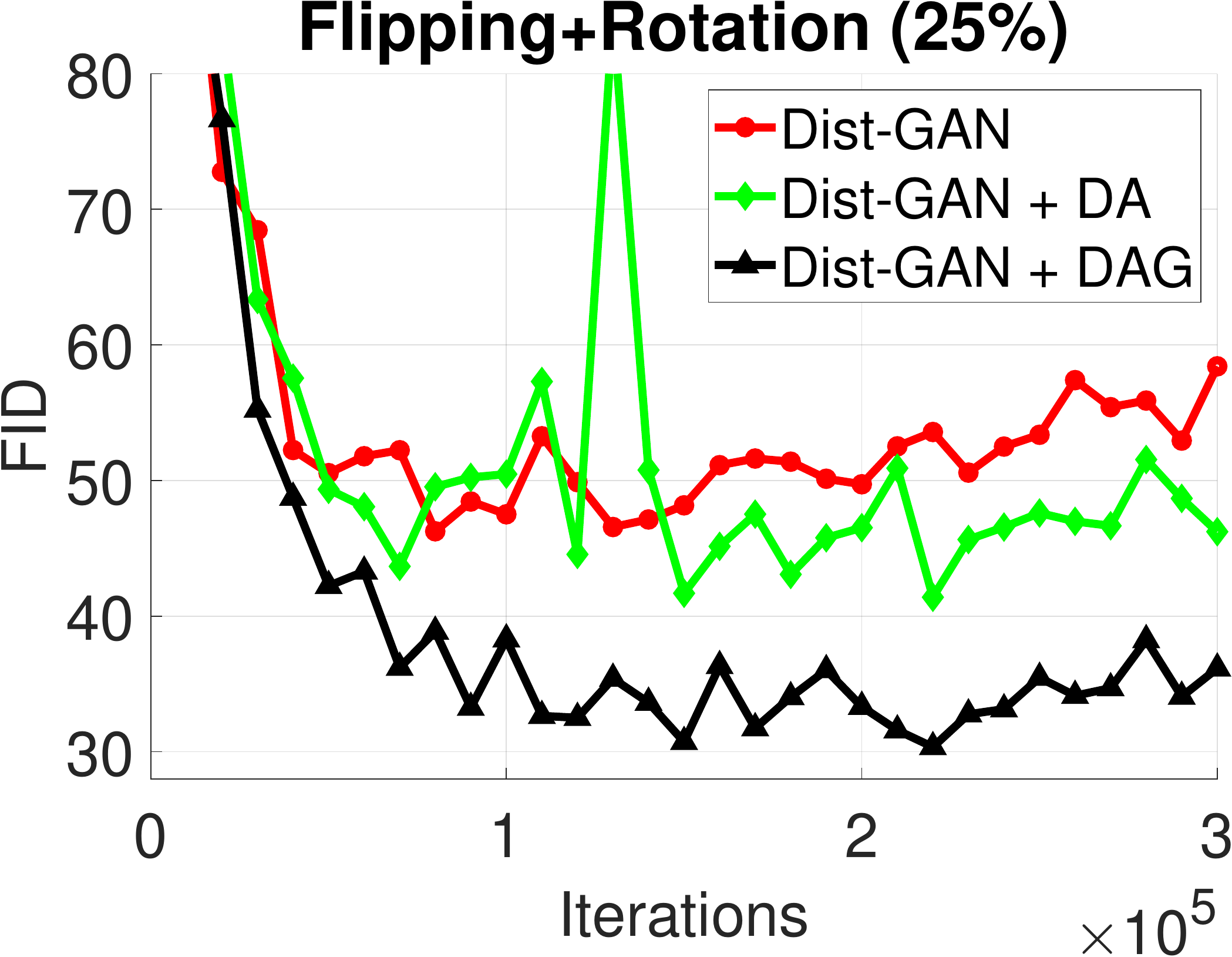}
  \caption{Comparing DA and our proposed DAG with SS-GAN \cite{chen-arxiv-2018} (first row) and Dist-GAN \cite{tran-eccv-2018} (second row) baselines on 25\% of dataset. Left to right columns: rotation, flipping, cropping, translation, and flipping+rotation. The horizontal axis is the number of training iterations, and the vertical axis is the FID score.}
  \label{data_argumentation_25}
\end{figure*}

In this study, experiments are conducted mainly on the CIFAR-10 dataset. We use small DC-GAN architecture (Refer to Appendix \ref{appendix_d} for details) to this study. We choose two state-of-the-art models: SS-GAN \cite{chen-arxiv-2018}, Dist-GAN \cite{tran-eccv-2018} as the baseline models. The common augmentation techniques in Table. \ref{da_techniques} are used in the experiment. In addition to the full dataset (100\%) of the CIFAR-10 dataset, we construct the subset with 25\% of CIFAR-10 dataset (randomly selected) as another dataset for our experiments. This small dataset is to investigate how the models address the problem of limited data. We compare \textbf{DAG} to \textbf{DA} and \textbf{Baseline}. \textbf{DA} is the classical way of applying GAN on the augmented dataset (similar to Section \ref{does_da_works_for_gan} of our toy example) and \textbf{Baseline} is training GAN models on the original dataset. Fig. \ref{data_argumentation_100} and Fig. \ref{data_argumentation_25} present the results on the full dataset (100\%) and 25\% of datasets respectively. Figures in the first row are with the SS-GAN, and figures in the second row are with the Dist-GAN. SS-GAN often diverges at about 100K iterations; therefore, we report its best FID within 100K. We summarize the best FID of these figures into Tables \ref{cifar_fid_ssgan}.

\begin{table*}

    \centering
    %\footnotesize
    %\scriptsize
    \caption{Best FID of SS-GAN (above) and Dist-GAN (below) baseline, DA and DAG methods on the CIFAR-10 dataset. FlipRot = Flipping + Rotation. We use K = 4 for all experiments (including FlipRot) as discussed in Table \ref{da_techniques} for a fair comparison.}
    \begin{tabular}{ c c c c c c c c c c c c}
    \toprule
    & & \multicolumn{2}{c}{\textbf{Rotation}} & \multicolumn{2}{c}{\textbf{Flipping}} & \multicolumn{2}{c}{\textbf{Cropping}} & \multicolumn{2}{c}{\textbf{Translation}} & \multicolumn{2}{c}{\textbf{FlipRot}}\\
    \cmidrule(r){3-4}  \cmidrule(r){5-6}  \cmidrule(r){7-8} \cmidrule(r){9-10} \cmidrule(r){11-12}
    \textbf{Data size} & \textbf{Baseline}  & \textbf{DA}     & \textbf{DAG}     & \textbf{DA}   & \textbf{DAG}  & \textbf{DA} & \textbf{DAG} & \textbf{DA} & \textbf{DAG} & \textbf{DA} & \textbf{DAG}\\
    \midrule
    \textbf{100\%} & 28.0 & 31.8 & 25.2 & 33.0  & 25.9 & 45.7 & 23.9 & 122.6 & 26.3 & 31.7 & 25.2 \\
    \textbf{25\%}  & 49.4 & 39.6 & 38.7    & 37.1 & 40.0 & 48.8 & 39.2 & 157.3 & 42.7 & 36.4 & 40.1 \\
    \midrule
    \midrule
    \textbf{100\%} & 29.6 & 49.0 & 23.7 & 40.1 & 25.0 & 55.3 & 24.2 & 134.6 & 25.5 & 42.1 & 23.3\\
    \textbf{25\%}  & 46.2 & 47.4 & 35.2 & 44.4 & 31.4 & 60.6 & 30.6 & 163.8 & 38.5 & 41.4 & 30.3\\
    \bottomrule
    \end{tabular}
    \label{cifar_fid_ssgan}

\end{table*}

First, we observe that applying DA for GAN does not support GAN to learn $P_d$ better than Baseline, despite few exceptions with SS-GAN on the 25\% dataset. Mostly, the distribution learned with DA is too different from the original one; therefore, the FIDs are often higher than those of the Baselines. In contrast, DAG improves the two Baseline models substantially with all augmentation techniques on both datasets.

Second, all of the augmentation techniques used with DAG improve both SS-GAN and Dist-GAN on two datasets. For 100\% dataset, the best improvement is with the Fliprot. For the 25\% dataset, Fliprot is competitive compared to other techniques. It is consistent with our theoretical analysis, and  invertible methods such as  Fliprot can provide consistent improvements. Note that, although cropping is non-invertible, utilizing this technique in our DAG still enables reasonable improvements from the Baseline. {\em This result further corroborates the effectiveness of our proposed framework using a range of data augmentation techniques, even non-invertible ones.}

Third, GAN becomes more fragile when training with fewer data, i.e., 25\% of the dataset. Specifically, on the full dataset GAN models converge stably, on the small dataset they both suffer divergence and mode collapse problems, especially SS-GAN. This is consistent with recent observations \cite{wang-eccv-2018transferring,brock-iclr-2018,donahue-arxiv-2019large,frid-neurocomputing-2018gan}: the more data GAN model trains on, the higher quality it can achieve. In the case of limited data, the performance gap between DAG versus DA and Baseline is even larger. Encouragingly, with only 25\% of the dataset, Dist-GAN + DAG with FlipRot still achieves similar FID scores as that of Baseline trained on the full dataset. DAG brings more significant improvements with Dist-GAN over SS-GAN. Therefore, we use Dist-GAN as the baseline in comparison with state of the art in the next section.

We also test our best version with limited data (10\% dataset). Our best DAG (K = 10, see Table \ref{distgan_on_k}) archives FID (=30.5) which is much better than baseline (=54.6) and comparable to the baseline on 100\% dataset (=29.6) in Table. \ref{cifar_fid_ssgan}.

\subsection{Comparison to state-of-the-art GAN}

\subsubsection{Self-supervised GAN + our proposed DAG}
In this section, we 
apply DAG (FlipRot augmentation) to  SS-DistGAN \cite{tran-nips-2019}, a self-supervised extension of DistGAN. We indicate this combination (SS-DistGAN + DAG) with FlipRot as our best system to compare to state-of-the-art methods. We also report (SS-DistGAN + DAG) with rotation to compare with previous works \cite{chen-arxiv-2018,tran-nips-2019} for fairness. We highlight the main results as follows.

\begin{table}

  %\footnotesize
  %\scriptsize
  \caption{FID scores with ResNet \cite{miyato-iclr-2018} on CIFAR-10 and STL-10 datasets. The FID scores are extracted from the respective papers when available. `*': 10K-10K FID is computed as in \cite{chen-arxiv-2018}. '+': 50K-50K FID is computed. All compared GANs are unconditional, except SAGAN and BigGAN. \textbf{R}: rotation and \textbf{F+R}: FlipRot.}
  \label{state_of_the_art}
  \centering
  \begin{tabular}{llll}
    \toprule
    \textbf{Methods} & \textbf{CIFAR-10}   & \textbf{STL-10}  & \textbf{CIFAR-10$ ^*$}\\
    \midrule
    SN-GAN \cite{miyato-iclr-2018}    			      & 21.70 $\pm$ .21 & 40.10 $\pm$ .50  & 19.73 \\
    SS-GAN \cite{chen-arxiv-2018}                     & -               & -                & 15.65 \\
    DistGAN \cite{tran-eccv-2018}  			          & 17.61 $\pm$ .30 & 28.50 $\pm$ .49  & 13.01 \\
    GN-GAN \cite{tran-aaai-2018}                      & 16.47 $\pm$ .28 & - & - \\
    %MSGAN$^+$ \cite{tran-nips-2019}                  & -  & -  & 19.89 \\
    MMD GAN$^+$ \cite{wang-arxiv-2018improving}       & - & 37.63$^+$ & 16.21$^+$  \\
    Auto-GAN$^+$ \cite{gong-iccv-2019autogan}         & - & 31.01$^+$ & 12.42$^+$  \\
    MS-DistGAN \cite{tran-nips-2019}                  & 13.90 $\pm$ .22 & 27.10 $\pm$ .34 & 11.40 \\
    \midrule
    SAGAN \cite{zhang-arxiv-2018} (cond.)             & 13.4  & - & - \\
    BigGAN \cite{brock-iclr-2018} (cond.)             & 14.73 & - & - \\
    \midrule
    %\textbf{SS-GAN$^+$}                               & - & - & - \\
    %\hline               
    %\textbf{Ours (SS-GAN$^+$)}                        & - & - & - \\
    \textbf{Ours (R)}                          & \textbf{13.72 $\pm$ .15}  & \textbf{25.69 $\pm$ .15} & \textbf{11.35} \\
    \textbf{Ours (F+R)}                        & \textbf{13.20 $\pm$ .19}  & \textbf{25.56 $\pm$ .15} & \textbf{10.89} \\
    \bottomrule
  \end{tabular}

\end{table}

We report our performance on natural images datasets: CIFAR-10, STL-10 (resized into $48 \times 48$ as in \cite{miyato-iclr-2018}). We investigate the performance of our best system. We use ResNet \cite{gulrajani-arxiv-2017,miyato-iclr-2018} (refer to Appendix \ref{appendix_d}) with ``hinge" loss as it attains better performance than standard ``log" loss \cite{miyato-iclr-2018}. We compare our proposed method to other state-of-the-art unconditional and conditional GANs. We emphasize that our proposed method is unconditional and does not use any labels.

Main results are shown in Table \ref{state_of_the_art}. The best FID attained in 300K iterations are reported as in \cite{xiang-arxiv-2017,li-nips-2017,tran-eccv-2018,yazici-arxiv-2018}. The ResNet is used for the comparison. We report our best system (SS-DistDAN + DAG) with Rotation and FlipRot. The improved performance over state-of-the-art GAN confirms the effectiveness of our proposed system.

In addition,
in Table \ref{state_of_the_art}, we also compare our FID to those of SAGAN \cite{zhang-arxiv-2018} and BigGAN \cite{brock-iclr-2018} (the current state-of-the-art conditional GANs). We perform the experiments under the same conditions using ResNet backbone on the CIFAR-10 dataset. The FID of SAGAN is extracted from \cite{tran-nips-2019}. For BigGAN, we extract the best FID from the original paper. Although our method does not use labeled data, our best FID approaches these state-of-the-art conditional GANs which use labeled data.
Our system
SS-DistDAN + DAG combines self-supervision as in \cite{tran-nips-2019} and optimized data augmentation to achieve outstanding performance.
Generated images using our system can be found in Figures \ref{fig:bestsystem} of Appendix \ref{appendix_b}. %Refer to Appendix \ref{implementation_computation} for discussion about the execution time of DAG.

%\noindent \textbf{Generated samples.} Fig. \ref{fig:bestsystem} shows the generated examples of our best system (SS-DistGAN + DAG with flipping+rotation). The left column is with the real samples, and the right column is with our generated samples. The first row is with CIFAR-10 and the second row is with STL-10. 

\begin{figure*}
    \centering
    \includegraphics[width=4.3cm,keepaspectratio]{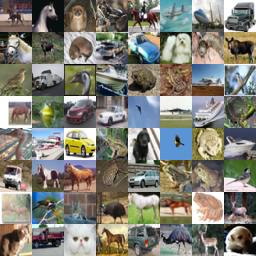}
    \includegraphics[width=4.3cm,keepaspectratio]{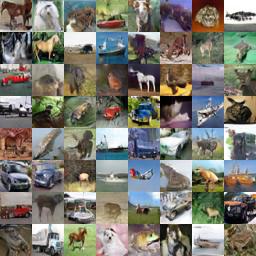}
    \includegraphics[width=4.3cm,keepaspectratio]{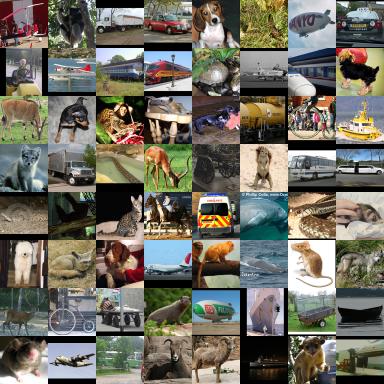}
    \includegraphics[width=4.3cm,keepaspectratio]{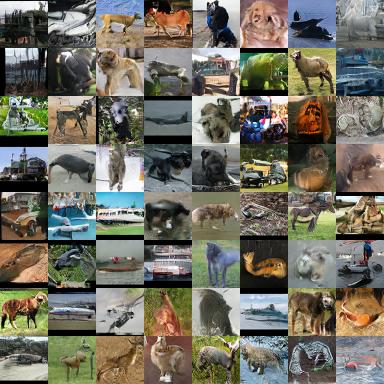}
    \caption{Real (left) and generated (right) examples by our best system on CIFAR-10 (two first columns) and STL-10 (two last columns).}
    \label{fig:bestsystem}
\end{figure*}

\subsubsection{Conditional GAN + our proposed DAG}
%\subsubsection{BigGAN + our proposed DAG and comparison with DiffAugment \cite{zhao-neurips-2020diffaugment}}
%We further demonstrate the effectiveness of the proposed DAG on BigGAN and conduct a comparison with DiffAugment \cite{zhao-neurips-2020diffaugment}, one of the recent work  concurrent with our work.
%We implement DAG on BigGAN using the published code of \cite{zhao-neurips-2020diffaugment}\footnote{https://github.com/mit-han-lab/data-efficient-gans}. 
We demonstrate that our DAG can also improve the state-of-the-art conditional GAN model, BigGAN \cite{brock-iclr-2018}. For this experiment, we apply rotations (0, 90, 180, 270 degrees) as transformations, and the model is trained with 60K iterations on limited datasets of CIFAR-10 (10\%, 20\%), CIFAR-100 (10\%, 20\%), and ImageNet (25\%). As shown in Table \ref{biggan_dag}, when DAG is applied to BigGAN, it can boost the performance of BigGAN considerably on these limited-size datasets. These experimental results for comparison were obtained with a single Titan RTX GPU with a batch size of 50. %The results of DiffAugment are reproduced from the published code.

% \begin{table*}
%   \centering
%   \caption{The IS and FID scores of BigGAN and  BigGAN + DAG on CIFAR-10, CIFAR-100, and ImageNet $128\times128$ with limited data (10\%, 20\%, and 25\% of original datasets). The augmentations of DAG used in this experiment are four rotations. The results of DiffAugment is reproduced from its published code.}
%   \begin{tabular}{l c c c c}
%     \toprule
%     \textbf{Dataset} & \multicolumn{2}{c}{BigGAN} & \multicolumn{2}{c}{BigGAN + DAG (rotation)}\\
%     \cmidrule(r){2-3}  \cmidrule(r){4-5}
%     & IS & FID & IS & FID \\
%     \hline
%     \hline
%     %\textbf{CIFAR-10 (100\%)} & 9.20  & 9.38  & 9.22 & 9.10\\
%     %\textbf{CIFAR-10 (20\%)}  & 8.87  & 22.75 & 8.98 & 17.61 \\
%     \textbf{CIFAR-10 (20\%)}   & 8.59  & 21.46 & 8.98 & 17.61 \\
%     \textbf{CIFAR-10 (10\%)}   & 7.23  & 45.82 & 7.87 & 36.85 \\
%     \textbf{CIFAR-100 (20\%)}  & 8.78  & 32.25 & 9.76 & 26.51 \\
%     \textbf{CIFAR-100 (10\%)}  & 6.71  & 68.54 & 7.50 & 51.14 \\    
%     \textbf{ImageNet (25\%)}   & 17.65 & 37.63 & 19.09 & 33.52 \\
%     \bottomrule
%   \end{tabular}
%   \label{biggan_dag}
% \end{table*}

\begin{table*}
  \centering
  \caption{The IS and FID scores of BigGAN and  BigGAN + DAG on CIFAR-10, CIFAR-100, and ImageNet $128\times128$ with limited data (10\%, 20\%, and 25\% of original datasets). The augmentations of DAG used in this experiment are four rotations. %The results of DiffAugment is reproduced from its published code. DiffAugment with translation and cutout as in the original paper.
  }
  \begin{tabular}{l c c c c c c c c c c}
    \toprule
    \textbf{Dataset} & \multicolumn{2}{c}{CIFAR-10 (20\%)} & \multicolumn{2}{c}{CIFAR-10 (10\%)} & \multicolumn{2}{c}{CIFAR-100 (20\%)} & \multicolumn{2}{c}{CIFAR-100 (10\%)} & \multicolumn{2}{c}{ImageNet (25\%)}\\
    \cmidrule(r){2-3}  \cmidrule(r){4-5} \cmidrule(r){6-7} \cmidrule(r){8-9} \cmidrule(r){10-11}
    \textbf{Scores} & IS & FID & IS & FID & IS & FID & IS & FID & IS & FID \\
    \hline
    \hline
    \textbf{BigGAN}                                                   & 8.51 & 22.3 & 7.03 & 48.3 & 8.78 & 32.25 & 6.71 & 68.54 & 17.65 & 37.63 \\
    %\textbf{BigGAN + DiffAugment (translation + cutout)} \cite{zhao-neurips-2020diffaugment} & 8.79 & 14.6 & 8.40 & 23.9 & 9.51 & 22.8 & 8.58 & 34.11     & -     & -     \\
    \textbf{BigGAN + DAG}                              & \textbf{8.98} & \textbf{17.6} & \textbf{7.87} & \textbf{36.9} & \textbf{9.76} & \textbf{26.51} & \textbf{7.50} & \textbf{51.14} & \textbf{19.09} & \textbf{33.52} \\
    %\textbf{BigGAN + our DAG (rotation + cropping)}                   & 8.83 & \textbf{14.1} & \textbf{8.63} & \textbf{23.6} & - & - & - & - & - \\
    \bottomrule
  \end{tabular}
  \label{biggan_dag}
\end{table*}

\subsubsection{Image-image translation + our proposed DAG} In this experiment, we apply DAG to CycleGAN \cite{zhu-cvpr-2017}, a GAN method for image-image translation. We use the public Pytorch code\footnote{https://github.com/junyanz/pytorch-CycleGAN-and-pix2pix} for our experiments. We follow the evaluation on Per-pixel accuracy, Per-class accuracy, and Class IOU as in \cite{zhu-cvpr-2017} on the Cityscapes dataset. Table \ref{tb:cyclegan} shows that CycleGAN + DAG achieves significantly better performance than the baseline for all three scores following the same setup. Note that we use DAG with only four rotations. We believe using more transformations will achieve more improvement. Note that (*) means the results reported in the original paper. The others are produced by the publicly available Pytorch code. The more detail implementation can be found in Appendix \ref{dag_cyclegan_impl}.

\begin{table}
  \centering
  \caption{FCN-scores for different methods, evaluated on Cityscapes labels $\rightarrow$ photo. We follow the exact setup as in CycleGAN \cite{zhu-cvpr-2017}. For all three scores, the higher is better. The augmentations of DAG in this experiment are four rotations. (*): The results reported in the original paper. The others are reproduced by the Pytorch code.}
  \begin{tabular}{l c c c}
    \toprule
    \textbf{Method} & \textbf{Per-pixel acc.} & \textbf{Per-class acc.} &  \textbf{Class IOU}   \\
    \hline
    \hline
    CycleGAN$(*)$  & 0.52 & 0.17 & 0.11 \\
    CycleGAN       & 0.21 & 0.06 & 0.02 \\
    CycleGAN + DAG & \textbf{0.59} & \textbf{0.19} & \textbf{0.15} \\
    \bottomrule
  \end{tabular}
  \label{tb:cyclegan}
\end{table}

\subsubsection{Mode collapse on Stacked MNIST}

We evaluate the stability of 
SS-DistGAN + DAG
%man-29dec:pls confirm above
%our best system 
and the diversity of its generator on Stacked MNIST \cite{metz-arxiv-2016}. Each image of this dataset is synthesized by stacking any three random MNIST digits. We follow the same setup with tiny architectures $K = \{\frac{1}{2}, \frac{1}{4}\}$ and evaluation protocol of \cite{metz-arxiv-2016}. $K$ indicates the size of the discriminator relative to the generator. We measure the quality of methods by the number of covered modes (higher is better) and KL divergence (lower is better) \cite{metz-arxiv-2016}. For this dataset, we report for our performance and compare to previous works as in Table. \ref{new_experiments_stacked_mnist}. The numbers show our proposed system outperforms the state of the art for both metrics. The results are computed from eight runs with the best parameters obtained via the same parameter as previous experiments. %Referring to Appendix \ref{appendix_b} for detail experiments.

\begin{table}

    \centering
    \scriptsize
    \caption{Comparing to state-of-the-art methods: Unrolled GAN \cite{metz-arxiv-2016}, WGAN-GP \cite{gulrajani-arxiv-2017}, Dist-GAN \cite{tran-eccv-2018}, Pro-GAN \cite{karras-iclr-2018}, MS-DistGAN  \cite{tran-nips-2019} on Stacked MNIST with tiny K=$\frac{1}{4}$ and K=$\frac{1}{2}$ architectures \cite{metz-arxiv-2016}. \textbf{R}: rotation and \textbf{F+R}: fliprot.}
    \begin{tabular}{ l  l  l  l  l  }
    \toprule
    & \multicolumn{2}{c}{\textbf{K=$\frac{1}{4}$}} & \multicolumn{2}{c}{\textbf{K=$\frac{1}{2}$}}\\
    \cmidrule(r){2-3}  \cmidrule(r){4-5}
    \textbf{Methods} &  \textbf{\#modes} & \textbf{KL} & \textbf{ \#modes} & \textbf{KL}\\
    \midrule
    \cite{metz-arxiv-2016} & 372.2 $\pm$ 20.7 & 4.66 $\pm$ 0.46 & 817.4 $\pm$ 39.9  & 1.43 $\pm$ 0.12 \\
    \cite{gulrajani-arxiv-2017} & 640.1 $\pm$ 136.3     & 1.97 $\pm$ 0.70 & 772.4 $\pm$ 146.5 & 1.35 $\pm$ 0.55  \\
    \cite{tran-eccv-2018} & 859.5 $\pm$ 68.7     & 1.04 $\pm$ 0.29 & 917.9 $\pm$ 69.6 & 1.06 $\pm$ 0.23 \\
    \cite{karras-iclr-2018} & 859.5 $\pm$ 36.2      & 1.05 $\pm$ 0.09 & 919.8 $\pm$ 35.1 & 0.82 $\pm$ 0.13 \\
    \cite{tran-nips-2019} & 926.7 $\pm$ 32.65 & 0.78 $\pm$ 0.13 & 976.0 $\pm$ 10.0 & 0.52 $\pm$ 0.07 \\
    \midrule
    \textbf{Ours (R)} & \textbf{947.4 $\pm$ 36.3} & \textbf{0.68 $\pm$ 0.14} & \textbf{983.7 $\pm$ 9.7} & \textbf{0.42 $\pm$ 0.11} \\
    \textbf{Ours (F+R)} & \textbf{972.9 $\pm$ 19.0} & \textbf{0.57 $\pm$ 0.12} & \textbf{981.5 $\pm$ 15.2} & \textbf{0.49 $\pm$ 0.15}\\
    \bottomrule
    \end{tabular}
    \label{new_experiments_stacked_mnist}
\end{table}

\subsection{Medical images with limited data}

We verify the effectiveness of our DAG on medical images with a limited number of samples. The experiment is conducted using the IXI dataset\footnote{https://brain-development.org/ixi-dataset/}, a public MRI dataset. In particular, we employ the T1 images of the HH subset (MRI of the brain). We extract two subsets: (i) 1000 images from 125 random subjects (8 slices per subject) (ii) 5024 images from 157 random subjects (32 slices per subject). All images are scaled to 64x64 pixels. We use DistGAN baseline with DCGAN architecture \cite{radford-arxiv-2015}, DAG with 90-rotation, and report the best FID scores. The results in Table \ref{distgan_on_medical_images} suggest that DAG improves the FID score of the baseline substantially and is much better than the baseline on the limited data. 

\begin{table}
  \footnotesize
  \caption{The experiments on medical images with a limited number of data samples. We use Dist-GAN as the baseline and report FID scores in this study.}
  \label{distgan_on_medical_images}
  \centering
  \begin{tabular}{ccc}
    \toprule
    \textbf{Data size} & \textbf{Baseline} & \textbf{Ours (Baseline + DAG)} \\
    \midrule
    1K samples & 71.12 & 46.83 \\
    \hline
    5K samples & 34.56 & 22.34 \\    
    \bottomrule
  \end{tabular}
\end{table}

\subsection{Training time comparison}
\label{implementation_computation}

Our GAN models are implemented with the Tensorflow deep learning framework \cite{tensorflow2015-whitepaper}. We measure the training time of DAG (K=4 branches) on our machine: Ubuntu 18.04, CPU Core i9, RAM 32GB, GPU GTX 1080Ti. We use DCGAN baseline (in Section \ref{exp_data_augmentation}) for the measurement. We compare models before and after incorporating DAG with SS-GAN and Dist-GAN. SS-GAN: 0.14 (s) per iteration. DistGAN: 0.11 (s) per iteration. After incorporating DAG, we have these training times: SS-GAN + DAG: 0.30 (s) per iteration and DistGAN-DAG:  0.23 (s) per iteration. The computation time is about $2\times$ higher with adding DAG (K = 4) and about $5\times$ higher with adding DAG (K = 10). Because of that, we propose to use K = 4 for most of the experiments which have a better trade-off between the FID scores and processing time and also is fair to compare to other methods. With K = 4, although the processing $2\times$ longer, DAG helps achieve good quality image generation, e.g. 25\% dataset + DAG has the same performance as 100\% dataset training, see our results of Dist-GAN + DAG with flipping+rotation. For most experiments in Section \ref{exp_data_augmentation}, we train our models on 8 cores of TPU v3 to speed up the training.
\section{Conclusion}

We propose a Data Augmentation optimized GAN (DAG) framework to improve GAN learning to capture the distribution of the original dataset. Our DAG can leverage the various data augmentation techniques to improve the learning stability of the discriminator and generator. We provide theoretical and empirical analysis to show that our DAG preserves the Jensen-Shannon (JS) divergence of original GAN with invertible transformations. Our theoretical and empirical analyses support the improved convergence of our design. Our proposed model can be easily incorporated into existing GAN models. Experimental results suggest that they help boost the performance of baselines implemented with various network architectures on the CIFAR-10, STL-10, and Stacked-MNIST datasets. The best version of our proposed method establishes state-of-the-art FID scores on all these benchmark datasets. Our method is applicable to address the limited data issue for GAN in many applications, e.g. medical applications.

% \fi
\IEEEpeerreviewmaketitle

\ifCLASSOPTIONcaptionsoff
  \newpage
\fi



\bibliographystyle{IEEEtran}
\bibliography{IEEEabrv,biblio}

\appendices

\section{}
\label{appendix_a}

\subsection{Proofs for theorems}
\label{proofs_for_theorems}

\noindent \textbf{Theorem \ref{js_theorem_1}} (Restate). Let $p_\mathbf{x}(\mathbf{x})$ and $q_\mathbf{x}(\mathbf{x})$ are two distributions in space $\mathbb{X}$. Let $T: \mathbb{X} \rightarrow \mathbb{Y}$ (linear or nonlinear) is differentiable and invertible mapping function (diffeomorphism) that transform $\mathbf{x}$ to $\mathbf{y}$. Under transformation $T$, distributions $p_\mathbf{x}(\mathbf{x})$ and $q_\mathbf{x}(\mathbf{x})$ are transformed to $p_\mathbf{y}(\mathbf{y})$ and $q_\mathbf{y}(\mathbf{y})$, respectively.
Therefore,
\begin{align}
d\mathbf{y} = |\mathcal{J}(\mathbf{x})|d\mathbf{x} \label{JSeq:1}\\
p_\mathbf{y}(\mathbf{y})=p_\mathbf{y}(T(\mathbf{x}))=p_\mathbf{x}(\mathbf{x})|\mathcal{J}(\mathbf{x})|^{-1} \label{JSeq:2}\\
q_\mathbf{y}(\mathbf{y})=q_\mathbf{y}(T(\mathbf{x}))=q_\mathbf{x}(\mathbf{x})|\mathcal{J}(\mathbf{x})|^{-1} \label{JSeq:3}
\end{align}
where $|\mathcal{J}(\mathbf{x})|$ is the determinant of the Jacobian matrix of $T$.
From \eqref{JSeq:2} and \eqref{JSeq:3}, we have:
\begin{equation}
p_\mathbf{y}(\mathbf{y})+q_\mathbf{y}(\mathbf{y})=p_\mathbf{x}(\mathbf{x})|\mathcal{J}(\mathbf{x})|^{-1}+q_\mathbf{x}(\mathbf{x})|\mathcal{J}(\mathbf{x})|^{-1} \label{JSeq:4}
\end{equation}
Let $m_\mathbf{y}=\frac{p_\mathbf{y}+q_\mathbf{y}}{2}$ and $m_\mathbf{x}=\frac{p_\mathbf{x}+q_\mathbf{x}}{2}$. From \eqref{JSeq:4}, we have equations:
\begin{equation}
\begin{split}
m_\mathbf{y}(\mathbf{y})&=\frac{p_\mathbf{y}(\mathbf{y})+q_\mathbf{y}(\mathbf{y})}{2}=\frac{p_\mathbf{x}(\mathbf{x})|\mathcal{J}(\mathbf{x})|^{-1}+q_\mathbf{x}(\mathbf{x})|\mathcal{J}(\mathbf{x})|^{-1}}{2}\\
&=\frac{p_\mathbf{x}(\mathbf{x})+q_\mathbf{x}(\mathbf{x})}{2}|\mathcal{J}(\mathbf{x})|^{-1} \label{JSeq:5}
\end{split}
\end{equation}
Since $m_\mathbf{x}(\mathbf{x})=\frac{p_\mathbf{x}(\mathbf{x})+q_\mathbf{x}(\mathbf{x})}{2}$, then,
\begin{equation}
m_\mathbf{y}(\mathbf{y}) = m_\mathbf{x}(\mathbf{x})|\mathcal{J}(\mathbf{x})|^{-1} \label{JSeq:6}
\end{equation}

\noindent From \eqref{JSeq:2}, \eqref{JSeq:3} and \eqref{JSeq:6}, we continue our proof as follows:

%\begin{figure*}
\begin{equation*}
\begin{split}
\mathrm{JS}(p_\mathbf{y}||q_\mathbf{y}) & = \frac{1}{2}\int \Big(p_\mathbf{y}(\mathbf{y})\log\big(\frac{p_\mathbf{y}(\mathbf{y})}{m_\mathbf{y}(\mathbf{y})}\big) \\
&+ q_\mathbf{y}(\mathbf{y})\log\big(\frac{q_\mathbf{y}(\mathbf{y})}{m_\mathbf{y}(\mathbf{y})}\big)\Big)d\mathbf{y}\\
&= \frac{1}{2}\int \Big(p_\mathbf{x}(\mathbf{x})|\mathcal{J}(\mathbf{x})|^{-1}\log\big(\frac{p_\mathbf{x}(\mathbf{x})|\mathcal{J}(\mathbf{x})|^{-1}}{m_\mathbf{y}(\mathbf{y})}\big) \\
&+ q_\mathbf{y}(\mathbf{y})\log\big(\frac{q_\mathbf{y}(\mathbf{y})}{m_\mathbf{y}(\mathbf{y})}\big)\Big)d\mathbf{y} \quad (\mathrm{from} \, \eqref{JSeq:2}) \\
& = \frac{1}{2}\int \Big(p_\mathbf{x}(\mathbf{x})|\mathcal{J}(\mathbf{x})|^{-1}\log(\frac{p_\mathbf{x}(\mathbf{x})|\mathcal{J}(\mathbf{x})|^{-1}}{m_\mathbf{y}(\mathbf{y})}) \\
&+
q_\mathbf{x}(\mathbf{x})|\mathcal{J}(\mathbf{x})|^{-1}\log(\frac{p_\mathbf{x}(\mathbf{x})|\mathcal{J}(\mathbf{x})|^{-1}}{m_\mathbf{y}(\mathbf{y})})\Big)d\mathbf{y}\\
& \quad (\mathrm{from} \, \eqref{JSeq:3}) \\
& = \frac{1}{2}\int |\mathcal{J}(\mathbf{x})|^{-1}\Big( p_\mathbf{x}(\mathbf{x})\log\big(\frac{p_\mathbf{x}(\mathbf{x})|\mathcal{J}(\mathbf{x})|^{-1}}{m_\mathbf{x}(\mathbf{x})|\mathcal{J}(\mathbf{x})|^{-1}}\big) \\
&+ q_\mathbf{x}(\mathbf{x})\log\big(\frac{q_\mathbf{x}(\mathbf{x})|\mathcal{J}(\mathbf{x})|^{-1}}{m_\mathbf{x}(\mathbf{x})|\mathcal{J}(\mathbf{x})|^{-1}}\big)\Big) d\mathbf{y}\\
& \quad (\mathrm{from} \, \eqref{JSeq:6})\\  
& = \frac{1}{2}\int |\mathcal{J}(\mathbf{x})|^{-1} \Big(p_\mathbf{x}(\mathbf{x})\log\big(\frac{p_\mathbf{x}(\mathbf{x})}{m_\mathbf{x}(\mathbf{x})|\mathcal{J}(\mathbf{x})|^{-1}}\big) \\
&+ q_\mathbf{x}(\mathbf{x})\log\big(\frac{q_\mathbf{x}(\mathbf{x})|\mathcal{J}(\mathbf{x})|^{-1}}{m_\mathbf{x}(\mathbf{x})|\mathcal{J}(\mathbf{x})|^{-1}}\big)\Big) |\mathcal{J}(\mathbf{x})|d\mathbf{x}\\
& \quad (\mathrm{from} \quad \eqref{JSeq:1})\\ 
& = \frac{1}{2}\int p_\mathbf{x}(\mathbf{x})\log(\frac{p_\mathbf{x}(\mathbf{x})}{m_\mathbf{x}(\mathbf{x})}) + q_\mathbf{x}(\mathbf{x})\log(\frac{q_\mathbf{x}(\mathbf{x})}{m_\mathbf{x}(\mathbf{x})})d\mathbf{x}\\
& =\mathrm{JS}(p_\mathbf{x} || q_\mathbf{x})
\end{split}
\label{js_proof_1}
\end{equation*}

% \begin{equation}
% \begin{split}
% \mathrm{JS}(p_\mathbf{y}||q_\mathbf{y}) & = \frac{1}{2}\int |\mathcal{J}(\mathbf{x})|^{-1} \Big(p_\mathbf{x}(\mathbf{x})\log\big(\frac{p_\mathbf{x}(\mathbf{x})}{m_\mathbf{x}(\mathbf{x})|\mathcal{J}(\mathbf{x})|^{-1}}\big) \\
% &+ q_\mathbf{x}(\mathbf{x})\log\big(\frac{q_\mathbf{x}(\mathbf{x})|\mathcal{J}(\mathbf{x})|^{-1}}{m_\mathbf{x}(\mathbf{x})|\mathcal{J}(\mathbf{x})|^{-1}}\big)\Big) |\mathcal{J}(\mathbf{x})|d\mathbf{x}\\
% & \quad (\mathrm{from} \quad \eqref{JSeq:1})\\ 
% & = \frac{1}{2}\int p_\mathbf{x}(\mathbf{x})\log(\frac{p_\mathbf{x}(\mathbf{x})}{m_\mathbf{x}(\mathbf{x})}) + q_\mathbf{x}(\mathbf{x})\log(\frac{q_\mathbf{x}(\mathbf{x})}{m_\mathbf{x}(\mathbf{x})})d\mathbf{x}\\
% & =\mathrm{JS}(p_\mathbf{x} || q_\mathbf{x})
% \end{split}
% \label{js_proof_2}
% \end{equation}

\noindent That concludes our proof. \\

\begin{lemma}
Let the sets of examples $\mathcal{X}^m$ have distributions $p^m$ respectively, $m=1,\dots,K$. Assume that the set $\mathcal{X}$ merges all samples of $\{\mathcal{X}^m\}$: $\mathcal{X} = \{\mathcal{X}^1, \dots, \mathcal{X}^K\}$ has the distribution $p$. Prove that the distribution $p$ can represented as the combination of distributions of its subsets: $p(\mathbf{x}) = \sum_{m=1}^K w_m p^m(\mathbf{x})$, $ \sum_{m=1}^K w_m = 1, w_m \geq 0$. \\
\label{lemma_1}
\end{lemma}

\noindent \textit{Proofs.} 

\noindent $\bullet$ The statement holds for $K = 1$, since we have: $p = p_1$. $w_1 = \sum_{m=1}^K w_m = 1$.

\noindent $\bullet$ For $K = 2$, let $\mathcal{X} = \{\mathcal{X}^1, \mathcal{X}^2\}$. We consider two cases:

\noindent a. If $\mathcal{X}^1$ and $\mathcal{X}^2$ are disjoint ($\mathcal{X}^1 \cap \mathcal{X}^2 = \emptyset$). Clearly, $p$ can be represented:

\begin{equation}
p(\mathbf{x}) = \underbrace{p(\mathbf{x} | \mathcal{X}^1)}_{w_1} p^1 + \underbrace{p(\mathbf{x} | \mathcal{X}^2)}_{w_2} p^2
\label{p_disjoint}    
\end{equation}

\noindent where $p(\mathbf{x} | \mathcal{X}^k)$ is the probability that $\mathbf{x} \in \mathcal{X}$ is from the subset $\mathcal{X}^k$, therefore $w_1 + w_2 = \sum_{m=1}^K w_m = 1$. The statement holds.

\noindent b. If $\mathcal{X}^1$ and $\mathcal{X}^2$ are intersection. Let $\mathcal{X}^1 \cap \mathcal{X}^2 = \mathcal{A}$. The set can be re-written: $\mathcal{X} = \{\underbrace{\mathcal{X}^{1-A}, \mathcal{A}}_{\mathcal{X}^1}, \underbrace{\mathcal{X}^{2-A}, \mathcal{A}}_{\mathcal{X}^2}\} = \{\underbrace{\mathcal{X}^{1-A},\mathcal{X}^{2-A}}_{\mathcal{X}^{12-A}},\mathcal{A},\mathcal{A}\}$, where $\mathcal{X}^{1-A} = \mathcal{X}^{1} \backslash A$ and $\mathcal{X}^{2-A} = \mathcal{X}^{2} \backslash A$. Since $\mathcal{X}^{12-A}$ (assume that it has its own distribution $p^{12-A}$) and $A$ (assume that it has its own distribution $p^A$) are disjoint, $p$ can be represented like Eq. \ref{p_disjoint}:

\begin{equation}
p(\mathbf{x}) = p(\mathbf{x} | \mathcal{X}^{12-A}) p^{12-A}(\mathbf{x}) + 2 p(\mathbf{x} | \mathcal{A}) p^A(\mathbf{x})
\label{p_intersect_1}   
\end{equation}

Note that since $\mathcal{X}^{1-A}$ (assume that it has distribution $p^{1-A}$) and $\mathcal{X}^{2-A}$ (assume that it has distribution $p^{2-A}$) are disjoint. Therefore, $ p^{12-A}$ can be written: $p^{12-A}(\mathbf{x}) = p^{12-A}(\mathbf{x} | \mathcal{X}^{1-A}) p^{1-A}(\mathbf{x}) + p^{12-A}(\mathbf{x} | \mathcal{X}^{2-A}) p^{2-A}(\mathbf{x})$. Substituting this into Eq. (\ref{p_intersect_1}), we have:

\begin{equation}
\begin{split}
p(\mathbf{x}) &= p(\mathbf{x} | \mathcal{X}^{12-A}) \big(p^{12-A}(\mathbf{x} | \mathcal{X}^{1-A}) p^{1-A}(\mathbf{x}) \\
&+ p^{12-A}(\mathbf{x} | \mathcal{X}^{2-A}) p^{2-A}(\mathbf{x}) \big) \\&+ 2 p(\mathbf{x} | \mathcal{A}) p^A(\mathbf{x})
\end{split}
\label{p_intersect_2}   
\end{equation}

Since two pairs ($\mathcal{X}^{1-A}$ and $A$) and ($\mathcal{X}^{2-A}$ and $A$) are also disjoint. Therefore, $p^1$ and $p^2$ can be represented:

\begin{equation}
p^1(\mathbf{x}) = p^1(\mathbf{x} | \mathcal{X}^{1-A}) p^{1-A}(\mathbf{x}) + p^1(\mathbf{x} | \mathcal{A}) p^A(\mathbf{x})
\label{p_intersect_3}   
\end{equation}

\begin{equation}
p^2(\mathbf{x}) = p^2(\mathbf{x} | \mathcal{X}^{2-A}) p^{2-A}(\mathbf{x}) + p^2(\mathbf{x} | \mathcal{A}) p^A(\mathbf{x})
\label{p_intersect_4}   
\end{equation}

Note that: 

\begin{equation}
\begin{split}
p(\mathbf{x} | \mathcal{X}^{12-A}) * p^{12-A}(\mathbf{x} | \mathcal{X}^{1-A}) &= p(\mathbf{x} | \mathcal{X}^{1-A}) \\&= p^1(\mathbf{x} | \mathcal{X}^{1-A}) * p(\mathbf{x} | X^1)
\end{split}
\label{p_intersect_5} 
\end{equation}

\begin{equation}
\begin{split}
p(\mathbf{x} | \mathcal{X}^{12-A}) * p^{12-A}(\mathbf{x} | \mathcal{X}^{2-A}) &= p(\mathbf{x} | \mathcal{X}^{2-A}) \\&= p^2(\mathbf{x} | \mathcal{X}^{2-A}) * p(\mathbf{x} | \mathcal{X}^2)
\end{split}
\label{p_intersect_6}
\end{equation}

From (\ref{p_intersect_3}), (\ref{p_intersect_4}), (\ref{p_intersect_5}), (\ref{p_intersect_6}), the Eq. (\ref{p_intersect_2}) is re-written:

\begin{equation}
p(\mathbf{x}) = \underbrace{p(\mathbf{x} | \mathcal{X}^1)}_{w_1} p^1(\mathbf{x}) + \underbrace{p(\mathbf{x} | \mathcal{X}^2)}_{w_2} p^2(\mathbf{x})
\end{equation}

The statement holds for $K$ = 2.\\

\noindent $\bullet$ Assume the statement holds with $K = k$, $k > 2$: $\mathcal{X} = \{\mathcal{X}^1, \dots, \mathcal{X}^k\}$ and $p = \sum_{m=1}^k w_m p^m$, $\sum_{m=1}^k w_m = 1$. We will prove the statement holds for $K = k + 1$.\\

\noindent Let $\mathcal{X} = \{\underbrace{\mathcal{X}^1, \dots, \mathcal{X}^k}_{\mathcal{X}^{1:k}}, \mathcal{X}^{k+1}\}$. Assume that $\mathcal{X}^{1:k}$ has distribution $p^{1:k}$ and $\mathcal{X}^{k+1}$ has distribution $p^{k+1}$. Thus,

\begin{equation}
\begin{split}
p(\mathbf{x}) &= (1 - w_{k+1})p^{1:k}(\mathbf{x}) + w_{k+1}p^{k+1}(\mathbf{x}) \\
     &= (1 - w_{k+1})\big(\sum_{m=1}^k w_m p^m(\mathbf{x})\big) + w_{k+1}p^{k+1}(\mathbf{x})\\
     &= \sum_{m=1}^{k+1} w'_m p^m(\mathbf{x})
\end{split}
\end{equation}

where $w'_m = (1 - w_{k+1}) * w_m, m \leq k$, and $w'_m = w_{k+1}, m = k + 1$. Clearly, $\sum_{m=1}^{k+1} w'_m = 1$. That concludes our proof.

\begin{lemma}
Considering two mixtures of distributions: $p = \sum_{m=1}^{K}w_{m}p^m$ and $q = \sum_{m=1}^{K}w_{m}q^m$. We have:
\begin{equation}
	\mathrm{JS}(p||q)\leq\sum_{m=1}^{K}w_{m}\mathrm{JS}(p^m||q^m)
\end{equation}
\label{lemma_2}
\end{lemma}

\noindent \textit{Proofs.} JS divergence is defined by: 

\begin{equation}
	\begin{split}
	& \mathrm{JS}(p||p) = \frac{1}{2}\mathrm{KL}(p||\frac{p+q}{2}) + \frac{1}{2}\mathrm{KL}(q||\frac{p+q}{2}) \\
	\end{split}\label{theorem2_js_1}
\end{equation}

From $p = \sum_{m=1}^{K}w_{m}p^m$ and $q = \sum_{m=1}^{K}w_{m}p^m$, we have:

\begin{equation}
	\begin{split}
	& \frac{p + q}{2} = \frac{\sum_{m=1}^{K}w_{m}p^m + \sum_{m=1}^{K}w_{m}q^m}{2} = \sum_{m=1}^{K}w_{m}\frac{(p^m+q^m)}{2}
	\end{split}\label{theorem2_js_2}
\end{equation}

Using the log-sum inequality: Given $ a_i \geq 0, b_i \geq 0, \forall i$, we have: $\sum_{m=1}^K a_i \log\frac{a_i}{b_i} \geq (\sum_{m=1}^K a_i) \log \frac{\sum_{m=1}^K a_i}{\sum_{m=1}^K b_i}$. We obtain the upper-bound of KL divergence as follows:

\begin{equation}
\begin{split}
	\mathrm{KL}(p||\frac{p+q}{2}) & = \mathrm{KL}(\sum_{m=1}^{K}w_{m}p^m||\sum_{m=1}^{K}w_{m}\frac{(p^m+q^m)}{2})\\
	&\leq \sum_{m=1}^{K}\mathrm{KL}(w_{m}p^m||w_{m}\frac{(p^m+q^m)}{2})
\end{split}\label{theorem2_kl_1}
\end{equation}

With equality if and only if $\frac{w_m p^m}{w_m p^m + w_m q^m} = \frac{p^m}{p^m + q^m}$ are equals for all $m$. Similarly, 
\begin{equation}
\begin{split}
	\mathrm{KL}(q||\frac{p+q}{2}) & = \mathrm{KL}(\sum_{m=1}^{K}w_{m}q^m||\sum_{m=1}^{K}w_{m}\frac{(p^m+q^m)}{2})\\
	&\leq \sum_{m=1}^{K}\mathrm{KL}(w_{m}q^m||w_{m}\frac{(p^m+q^m)}{2})
\end{split}\label{theorem2_kl_2}
\end{equation}

From Eqs \ref{theorem2_js_1}), (\ref{theorem2_kl_1}), and (\ref{theorem2_kl_2}), we have: 

\begin{equation}
	\begin{split}
	& \mathrm{JS}(p||q) \\
	& \leq \frac{1}{2}\sum_{m=1}^{K}\mathrm{KL}(w_{m}p^m||w_{m}\frac{(p^m+q^m)}{2})\\ &+\frac{1}{2}\sum_{m=1}^{K}\mathrm{KL}(w_{m}q^m||w_{m}\frac{(p^m+q^m)}{2}) \\
	& = \frac{1}{2}\sum_{m=1}^{K}\mathrm{KL}(w_{m}p^m||w_{m}\frac{(p^m+q^m)}{2}) \\
	&+ \mathrm{KL}(w_{m}q^m||w_{m}\frac{(p^m+q^m)}{2})\\
	& = \sum_{m=1}^{K}\mathrm{JS}(w_{m}p^m||w_{m}q^m) = \sum_{m=1}^{K}w_{m}\mathrm{JS}(p^m||q^m)
	\end{split}
\end{equation}

\noindent That concludes our proof.

\subsection{Implementation details}

\subsubsection{The implementation of DAG for GAN models}

In our implementation of DAG, we compute the average of K branches (including the identity branch) for DAG as shown in Eqs. \ref{gan_dis_aux_impl}, \ref{gan_gen_aux_impl} instead of K - 1 branches, which we found empirically more stable for most of our GAN models. The parameters $\lambda_u, \lambda_v$ are tuned in our experiments according to these objectives.
 
\begin{equation}
\begin{split}
\max_{D,\{D_k\}}\mathcal{V}(D,&\{D_k\},G) = \mathcal{V}(D,G) + \frac{\lambda_u}{K}\sum_{k=1}^{K} \mathcal{V}(D_k,G)
\end{split}
\label{gan_dis_aux_impl}
\end{equation}

\begin{equation}
\begin{split}
\min_{G}\mathcal{V}(D,&\{D_k\},G) = \mathcal{V}(D,G) + \frac{\lambda_v}{K}\sum_{k=1}^{K} \mathcal{V}(D_k,G)
\end{split}
\label{gan_gen_aux_impl}
\end{equation}

where $D = D_1$. \\

\subsubsection{The implementation of DAG for CycleGAN}
\label{dag_cyclegan_impl}

We apply DAG for adversarial losses of CycleGAN \cite{zhu-cvpr-2017} as following: 

\begin{equation}
\begin{split}
\mathcal{L}_{\mathrm{GAN}}(G, D_Y^k, X, Y) &= \mathbb{E}_{\mathbf{y} \sim {P_d(\mathbf{y})^{T_k}}}\log \Big(D_Y^k(\mathbf{y})\Big) \\
&+ \mathbb{E}_{\mathbf{x} \sim {P_d(\mathbf{x})^{T_k}}}\log \Big(1-D_Y^k(\mathbf{x})\Big) \end{split}
\end{equation}

where G is the mapping function (generator) $G: X \rightarrow Y$ and $D_Y$ is the discriminator on domain $Y$. $D_Y^k$ is the discriminator on samples transformed by the transformer $T_k$, e.g., $Y$ (distribution $P_d(\mathbf{y})$) is transformed into $Y^k$ (distribution $P_d(\mathbf{y})^{T_k}$). Similarly for the mapping function $F: Y \rightarrow X$ and and its discriminator $D_X$ as well: e.g., $\mathcal{L}_{GAN}(F, D_X^k, Y, X)$. The objectives of CycleGAN + DAG are written:

\begin{equation}
\begin{split}
& \min_{G,F} \max_{D_X, D_Y} \mathcal{L}(G, F, D_X, D_Y) \\
& = \mathcal{L}_{\mathrm{GAN}}(G, D_Y, X, Y) + \frac{\lambda_v}{K}\sum_{k=1}^{K} \mathcal{L}_{GAN}(G, D_Y^k, X, Y) \\
& + \mathcal{L}_{\mathrm{GAN}}(F, D_X, Y, X) + \frac{\lambda_v}{K}\sum_{k=1}^{K} \mathcal{L}_{GAN}(F, D_X^k, Y, X) \\
& + \mathcal{L}_{\mathrm{cyc}}(G, F)
\end{split}
\end{equation}

Here, $D_X^1 = D_X$ and $D_Y^1 = D_Y$. $\mathcal{L}_{\mathrm{cyc}}(G, F)$ is the same as in the original paper. The discriminators $D_Y^k$ shares weights except the last layers, similarly for discriminators $D_X^k$.\\

\section{}
\label{appendix_b}

\subsection{DAG model diagrams}

\begin{figure}
    \centering
    \includegraphics[width=8cm,keepaspectratio]{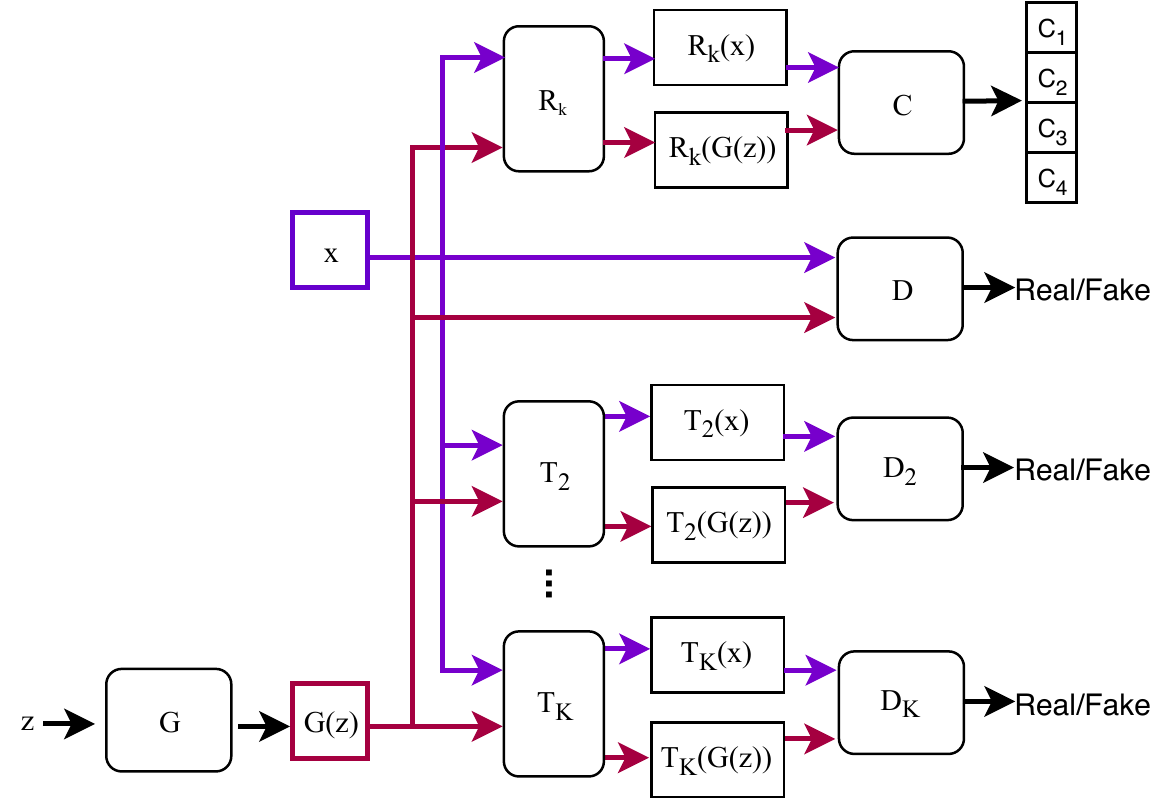}
    \caption{Applying our DAG for SSGAN model. R$_k, k=1\dots4$ are the rotation techniques (0$^\circ$, 90$^\circ$, 180$^\circ$, 270$^\circ$) and the classifier $C$ used in the self-supervised task of the original SSGAN. Refer to \cite{chen-arxiv-2018} for details of SSGAN. We apply T$_k, k=1 \dots K$ as the augmentation techniques for our DAG. Note that the DAG paths (bottom-right) are in fact the same as that for the vanilla GAN as shown in Figure~\ref{dag_models} of our main paper: DAG involves a stack of real/fake discriminators for transformed samples. This shows that the same DAG design is generally applicable to other GAN models.}
    \label{fig:framework-ssgan-dag}
\end{figure}

\begin{figure}
    \centering
    \includegraphics[width=8cm,keepaspectratio]{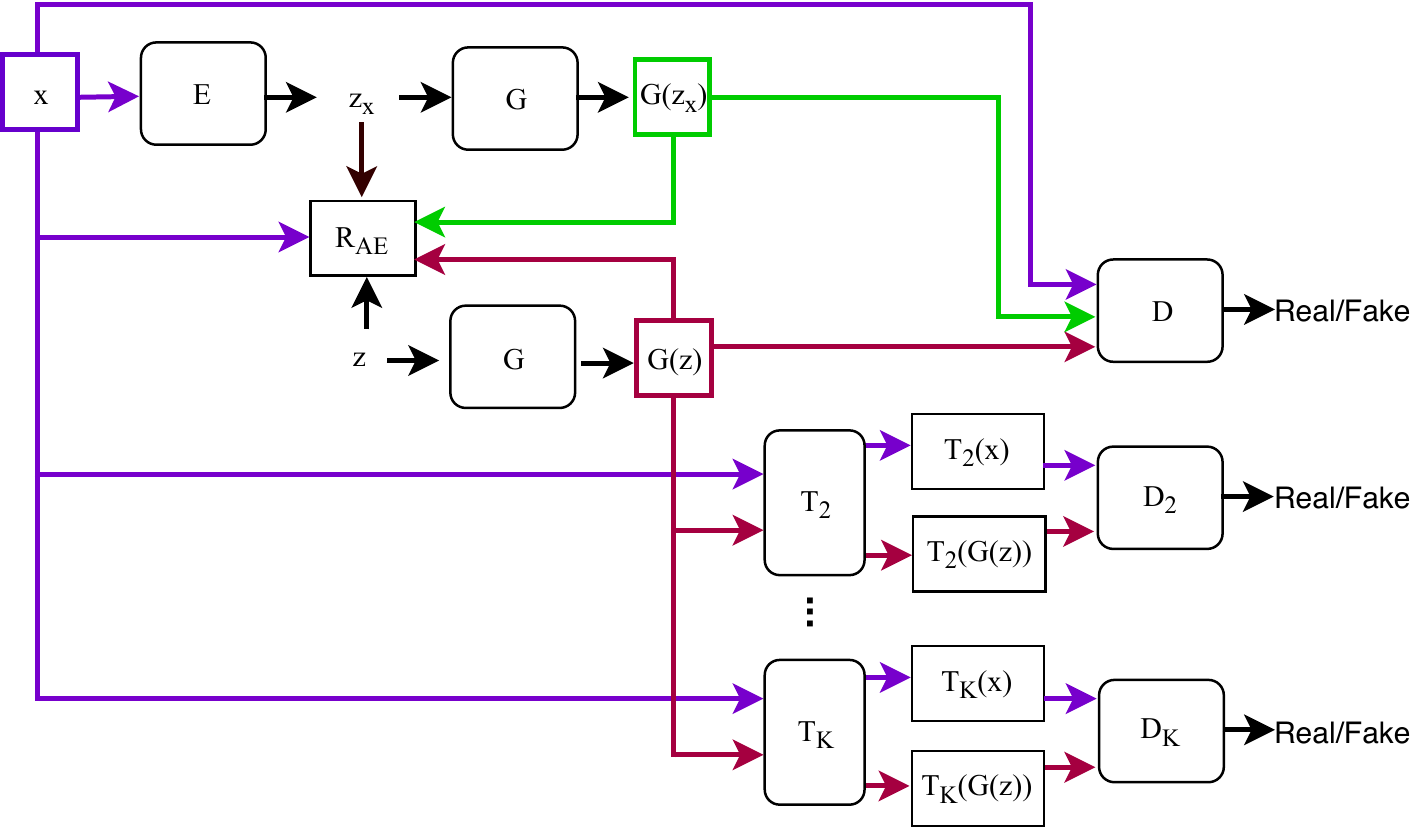}
    \caption{Our DAG  applied for DistGAN model (Refer to \cite{tran-eccv-2018} for the details). Here, we emphasize the difference is the DAG with T$_k$ branches. T$_k$ are the augmentation techniques used in our DAG. Furthermore, we note that the DAG paths (bottom-right) are in fact the same as that for the vanilla GAN as shown in Figure~\ref{dag_models} of our main paper: DAG involves a stack of real/fake discriminators for transformed samples. This shows that the same DAG design is generally applicable to other GAN models.}
    \label{fig:framework-distgan-dag}
\end{figure}

Figures \ref{fig:framework-ssgan-dag} and \ref{fig:framework-distgan-dag} present model diagrams of applying DAG for DistGAN and SSGAN baselines respectively. We keep the components of the original baseline models and only apply our DAG with branches of T$_k$ for the generators and discriminators.
From these diagrams, it is clear that the DAG paths are the same as that for the vanilla GAN as shown in Figure~\ref{dag_models} of our main paper: DAG involves a stack of real/fake discriminators for transformed samples. These examples show that the same DAG design is generally applicable to other GAN models.\\

\subsection{Generated examples on MNIST dataset}

\begin{figure}
  \centering
  
    \includegraphics[width=2.8cm,keepaspectratio]{image_100000_real_from_rotation.jpg}
    \includegraphics[width=2.8cm,keepaspectratio]{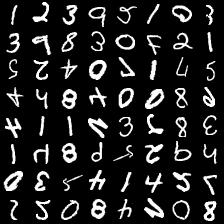}
    \includegraphics[width=2.8cm,keepaspectratio]{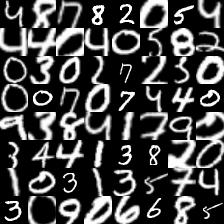}
    
    \vspace{0.1cm}
    
    \includegraphics[width=2.8cm,keepaspectratio]{gan_baseline_200000_fake.jpg}
    \includegraphics[width=2.8cm,keepaspectratio]{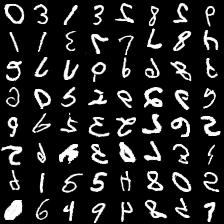}
    \includegraphics[width=2.8cm,keepaspectratio]{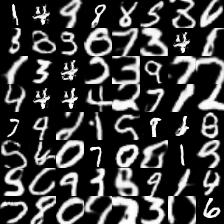}
    
  \caption{The generated examples of toy experiment on the full dataset (100\%). First rows: the real samples and real augmented samples. Second rows: generated samples. First column: the real samples, the generated samples of the GAN baseline. Second column: flipped real samples,  and the generated samples of DA with flipping. Third column: the cropped real samples and the generated samples of DA with cropping.}
  \label{mnist_data_agumentation_others}
\end{figure}

Figure \ref{mnist_data_agumentation_others} shows more generated samples for the toy example with DA methods (flipping and cropping) on MNIST dataset (please refer to Section~\ref{does_da_works_for_gan}
of our main paper). In the first column is with the real samples and the generated samples of the Baseline. The second column is with flipped real samples and the generated samples of DA with flipping. The last column is with the cropped real samples and the generated samples of DA with cropping.

\subsection{Augmented examples on CIFAR-10 and STL-10 datasets}

\begin{figure}
    \centering
    % real
    \includegraphics[width=2.8cm,keepaspectratio]{figures/image_0_real.jpg}
    % rotation
    \includegraphics[width=2.8cm,keepaspectratio]{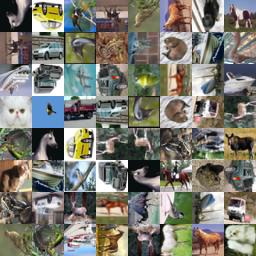}
    % flipping
     \includegraphics[width=2.8cm,keepaspectratio]{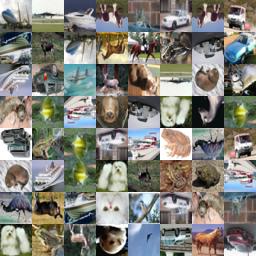}
     
    \vspace{0.1cm}
    
    % translation
    \includegraphics[width=2.8cm,keepaspectratio]{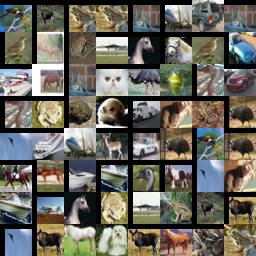}
    % flipping + rotation
    \includegraphics[width=2.8cm,keepaspectratio]{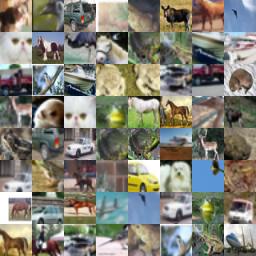}    
    % flipping + rotation
    \includegraphics[width=2.8cm,keepaspectratio]{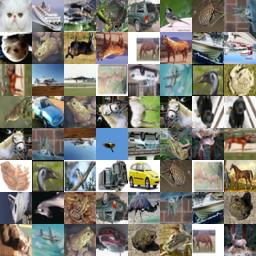}
    %\includegraphics[width=7cm,keepaspectratio]{figures/cifar10_dcgan_distgan_dag_flipping_100_270000.jpg}
    % translation
    %\includegraphics[width=7cm,keepaspectratio]{figures/cifar10_dcgan_distgan_dag_translation_100_300000.jpg}
    % cropping
    %\includegraphics[width=7cm,keepaspectratio]{figures/cifar10_dcgan_distgan_dag_cropping_100_300000.jpg}
    % flipping+rotation
    \caption{Examples of real and transformed real samples of CIFAR-10 used to train DA and DAG. Figures from left to right and top to bottom: the real samples, the rotated real samples, the flipped real samples, the translated real samples, the cropped real samples, and the flipped+rotated real samples.}
    \label{fig:cifar10-da-dag-100}
\end{figure}

\noindent \textbf{Examples of transformed real samples.} Figure \ref{fig:cifar10-da-dag-100} illustrates examples of transformed real samples we used to augment our training CIFAR-10 dataset. From left to right and top to bottom are with the original real samples, the rotated real samples, the flipped real samples, the translated real samples, the cropped real samples, and the flipped+rotated real samples.\\

\section{}
\label{appendix_d}

\subsection{DCGAN Networks}

We use the small DCGAN backbone for the study of data augmentation on CIFAR-10. Our DCGAN networks are presented in Table. \ref{dcgan} for the encoder, the generator, and the discriminator.

\begin{table*}[ht!]
	\caption{\label{dcgan}Our DCGAN architecture is similar to \cite{radford-arxiv-2015} but the smaller number of feature maps (D = 64) to be more efficient for our ablation study on CIFAR-10. The Encoder is the mirror of the Generator. Slopes of lReLU functions are set to $0.2$. $\mathcal{U}(0, 1)$ is the uniform distribution. $M = 32$. Discriminator for CIFAR-10: three different heads for GAN task and auxiliary tasks. K = 4 in our implementation.}
   	\centering
   	\scriptsize
    \begin{minipage}{.3\linewidth}
    	\centering
    	{\begin{tabular}{c}
			\toprule
			\midrule
		 	RGB image $x\in \bbR^{M\times M \times 3}$ \\
            \midrule
            5$\times$5, stride=2 conv. 1 $\times$ D ReLU\\
            \midrule
            5$\times$5, stride=2 conv. BN 2 $\times$ D ReLU\\        		 	
			\midrule
            5$\times$5, stride=2 conv. BN 4 $\times$ D ReLU\\            	
            \midrule
            5$\times$5, stride=2 conv. BN 8 $\times$ D ReLU\\    
            \midrule          
            dense $\rightarrow$ 128 \\
            \midrule
			\bottomrule
			\midrule
			\textbf{Encoder for CIFAR-10}\\
		\end{tabular}}
    \end{minipage}   	
    \begin{minipage}{.3\linewidth}
    	\centering
    	{\begin{tabular}{c}
			\toprule
			\midrule
		 	$z\in \bbR^{128} \sim \mathcal{U}(0, 1)$ \\	 	
           	\midrule
            dense $\rightarrow$ 2 $\times$ 2 $\times$ 8 $\times$ D  \\
            \midrule
            5$\times$5, stride=2 deconv. BN 4 $\times$ D ReLU\\
			\midrule
            5$\times$5, stride=2 deconv. BN 2 $\times$ D ReLU\\
            \midrule
            5$\times$5, stride=2 deconv. BN 1 $\times$ D ReLU\\
            \midrule
            5$\times$5, stride=2 deconv. 3 Sigmoid\\	
            \midrule
			\bottomrule
			\midrule
			\textbf{Generator for CIFAR-10}\\
		\end{tabular}}
    \end{minipage}
    \begin{minipage}{.3\linewidth}
    	\centering
    	{\begin{tabular}{c}
			\toprule
			\midrule
			RGB image $x\in \bbR^{M\times M \times 3}$ \\
            \midrule
            5$\times$5, stride=2 conv. 1 $\times$ D lReLU\\
            \midrule
            5$\times$5, stride=2 conv. BN 2 $\times$ D lReLU\\        		 	
			\midrule
            5$\times$5, stride=2 conv. BN 4 $\times$ D lReLU\\            	
            \midrule
            5$\times$5, stride=2 conv. BN 8 $\times$ D lReLU\\    
            \midrule
            dense $\rightarrow$ 1 (GAN task)\\
            dense $\rightarrow$ K - 1 (K - 1 augmented GAN tasks) \\
            \midrule
			\bottomrule
		    \midrule
			\textbf{Discriminator for CIFAR-10}\\
		\end{tabular}}
    \end{minipage}
\end{table*}

\subsection{Residual Networks}

Our Residual Networks (ResNet) backbones of the encoders, the generators and the discriminators for CIFAR-10 and STL-10 datasets are presented in Table. \ref{tab:resnets_cifar10} and Table. \ref{tab:resnets_stl} respectively (the same as in \cite{miyato-iclr-2018}).

\begin{figure*}[ht!]
	\begin{tabular}{cc}
	     \scriptsize
        \begin{minipage}{1.\textwidth}
          \tblcaption{\label{tab:resnets_cifar10}ResNet architecture for CIFAR10 dataset. The Encoder is the mirror of the Generator. We use similar architectures and ResBlock to the ones used in \cite{miyato-iclr-2018}. $\mathcal{U}(0, 1)$ is the uniform distribution. Discriminator. K different heads for GAN task and auxiliary tasks. K = 4 in our implementation.}
          \centering
          \begin{minipage}{.25\textwidth}
                        \centering
                        {\begin{tabular}{c}
                            \toprule
                            \midrule
                            RGB image $x\in \bbR^{32\times 32 \times 3}$ \\
                            \midrule
                            3$\times$3 stride=1, conv. 256\\ 
                            \midrule
                            ResBlock down 256\\                
                            \midrule
                            ResBlock down 256\\
                            \midrule
                            ResBlock down 256\\                                                         
                            \midrule
                            dense $\rightarrow$ 128 \\
                            \midrule
                            \bottomrule
                            \midrule
                            \textbf{Encoder for CIFAR}
                        \end{tabular}}
                    \end{minipage}
          \begin{minipage}{.25\textwidth}
              \centering
              {\begin{tabular}{c}
                  \toprule
                  \midrule
                  $z\in \bbR^{128} \sim \mathcal{U}(0, 1)$ \\
                  \midrule
                  dense, $4 \times 4 \times 256$ \\
                  \midrule
                  ResBlock up 256\\
                  \midrule
                  ResBlock up 256\\
                  \midrule
                  ResBlock up 256\\
                  \midrule
                  BN, ReLU, 3$\times$3 conv, 3 Sigmoid\\
                  \midrule
                  \bottomrule
                  \midrule
                  \textbf{Generator for CIFAR}\\
              \end{tabular}}
          \end{minipage}
          \begin{minipage}{.4\textwidth}
              \centering
              {\begin{tabular}{c}
                  \toprule
                  \midrule
                  RGB image $x\in \bbR^{32\times 32 \times 3}$ \\
                  \midrule
                  ResBlock down 128\\
                  \midrule
                  ResBlock down 128\\
                  \midrule
                  ResBlock 128\\
                  \midrule
                  ResBlock 128\\
                  \midrule
                  ReLU\\
                  \midrule
		          dense $\rightarrow$ 1 (GAN task)\\
		          dense $\rightarrow$ K - 1 (K - 1 augmented GAN tasks) \\
                  \midrule
                  \bottomrule
                  \midrule
                  \textbf{Discriminator for CIFAR}\\
              \end{tabular}}
          \end{minipage}
        \end{minipage}
    \end{tabular}
\end{figure*}

\begin{table*}[ht!]
          \caption{\label{tab:resnets_stl}ResNet architecture for STL-10 dataset. The Encoder is the mirror of the Generator. We use similar architectures and ResBlock to the ones used in \cite{miyato-iclr-2018}. $\mathcal{U}(0, 1)$ is the uniform distribution. For discriminator, different heads for GAN task and auxiliary tasks. K = 4 in our implementation.}
          \centering
          \scriptsize
          \begin{minipage}{.3\textwidth}
                        \centering
                        {\begin{tabular}{c}
                            \toprule
                            \midrule
                            RGB image $x\in \bbR^{48\times 48 \times 3}$\\
                            \midrule
                            3$\times$3 stride=1, conv. 64\\                            
                            \midrule
                            ResBlock down 128\\
                            \midrule 
                            ResBlock down 256\\
                            \midrule 
                            ResBlock down 512\\
                            \midrule                                 
                            dense $\rightarrow$ 128 \\
                            \midrule
                            \bottomrule
                            \midrule
                            \textbf{Encoder for STL-10}\\
                        \end{tabular}}
                        
                    \end{minipage}
          \begin{minipage}{.3\textwidth}
              \centering
              {\begin{tabular}{c}
                  \toprule
                  \midrule
                  $z\in \bbR^{128} \sim \mathcal{U}(0, 1)$ \\
                  \midrule
                  dense, $6 \times 6 \times 512$ \\
                  \midrule
                  ResBlock up 256\\
                  \midrule
                  ResBlock up 128\\
                  \midrule
                  ResBlock up 64\\
                  \midrule
                  BN, ReLU, 3$\times$3 conv, 3 Sigmoid\\
                  \midrule
                  \bottomrule
                  \midrule
                  \textbf{Generator for STL-10}\\
              \end{tabular}}
          \end{minipage}
          \begin{minipage}{.3\textwidth}
              \centering
              {\begin{tabular}{c}
                  \toprule
                  \midrule
                  RGB image $x\in \bbR^{48\times 48 \times 3}$ \\
                  \midrule
                  ResBlock down 64\\
                  \midrule
                  ResBlock down 128\\
                  \midrule
                  ResBlock down 256\\
                  \midrule
                  ResBlock down 512\\
                  \midrule
                  ResBlock 1024\\
                  \midrule
                  ReLU\\
                  \midrule
		          dense $\rightarrow$ 1 (GAN task)\\
		          dense $\rightarrow$ K - 1 (K - 1 augmented GAN tasks) \\
                  \midrule
                  \bottomrule
                  \midrule
                  \textbf{Discriminator for STL-10}
              \end{tabular}}
          \end{minipage}
\end{table*}

\end{document}